\documentclass{article} 
\usepackage[preprint]{neurips_2021}

\usepackage{amsmath,amsfonts,bm}

\def\eqref#1{equation~\ref{#1}}

\DeclareMathAlphabet{\mathsfit}{\encodingdefault}{\sfdefault}{m}{sl}
\SetMathAlphabet{\mathsfit}{bold}{\encodingdefault}{\sfdefault}{bx}{n}

\usepackage{url}
\usepackage{graphicx}
\usepackage{subcaption}
\usepackage{outlines}
\usepackage{comment}
\usepackage[title]{appendix}
\usepackage[utf8]{inputenc} 
\usepackage[T1]{fontenc}    
\usepackage[draft]{hyperref}       
\usepackage{url}            
\usepackage{amsmath,amsfonts,amssymb}       
\usepackage{bm}       
\usepackage{nicefrac}       
\usepackage{microtype}      
\usepackage[dvipsnames,table]{xcolor}
\usepackage{scalerel}
\usepackage[capitalise,nameinlink]{cleveref}
\usepackage{wrapfig}
\usepackage{booktabs}
\usepackage{algorithm}
\usepackage[noend]{algpseudocode}
\usepackage[normalem]{ulem}
\usepackage{todonotes}
\usepackage{authblk}
\usepackage{asciilist}

\DeclareMathOperator*{\E}{\scaleobj{1.1}{\mathbb{E}}}

\newcommand{\qex}{Q_{\text{explore}}}

\newcommand{\tex}{\tau_{\text{explore}}}

\newcommand{\pitask}{\pi_{\text{task}}}
\newcommand{\piexplore}{\pi_{\text{explore}}}
\newcommand{\mdp}{\mathcal{M}}

\definecolor{niceblue}{HTML}{2F80ED}

\hypersetup{
    colorlinks=true,
    linkcolor=niceblue,
    filecolor=magenta,
    urlcolor=niceblue,
    citecolor=black,
}

\newcommand{\algname}{Decoupled Exploration and Exploitation Policies} 
\newcommand{\algshort}{DEEP} 

\title{Decoupled Exploration and Exploitation Policies \\for Sample-Efficient Reinforcement Learning}

\author[1]{William F. Whitney}
\author[2]{Michael Bloesch}
\author[2]{Jost Tobias Springenberg}
\author[2]{Abbas Abdolmaleki}
\author[1]{Kyunghyun Cho}
\author[2]{Martin Riedmiller}
\affil[1]{Department of Computer Science, New York University}
\affil[2]{DeepMind, London}

\begin{document}

\maketitle

\begin{abstract}
Despite the close connection between exploration and sample efficiency, most state of the art reinforcement learning algorithms 
include no considerations for exploration beyond maximizing the entropy of the policy.
In this work we address this seeming missed opportunity.
We observe that the most common formulation of directed exploration in deep RL, known as bonus-based exploration (BBE), suffers from bias and slow coverage in the few-sample regime.
This causes BBE to be actively detrimental to policy learning in many control tasks.
We show that by decoupling the task policy from the exploration policy, directed exploration can be highly effective for sample-efficient continuous control.
Our method, \algname{} (\algshort{}), can be combined with any off-policy RL algorithm without modification.
When used in conjunction with soft actor-critic, \algshort{} incurs no performance penalty in densely-rewarding environments.
On sparse environments, \algshort{} gives a several-fold improvement in data efficiency due to better exploration.
\end{abstract}

\section{Introduction}
Recent progress in reinforcement learning (RL) for continuous control has led to significant improvements in sample complexity and performance.
While earlier on-policy algorithms required hundreds of millions of environment steps to learn, recent off-policy algorithms have brought the sample complexity of model-free RL in range of solving tasks on real robots \citep{haarnoja2018softB}.

In parallel, a rich literature has been developed for directed exploration in deep reinforcement learning, inspired in part by the theoretical impact of exploration on sample complexity.
The bulk of these methods fall into the family of bonus-based exploration (BBE) methods, in which a policy receives a bonus for visiting states deemed to be interesting or novel.
BBE algorithms have enabled RL to solve a variety of long-horizon, sparse-reward tasks, most notably the game Montezuma's Revenge from the Arcade Learning Environment (ALE) \citep{Bellemare2015TheAL}.

These two subfields both aim to minimize the sample complexity of model-free RL, and their methods are in principle perfectly complementary.
Off-policy algorithms extract improved policies from data collected by (notionally) arbitrary behavior, and their performance is limited only by the coverage of the data;
meanwhile exploration generates data with improved coverage.
However, to date the impact of directed exploration techniques on sample-efficient control has been minimal, with state of the art algorithms using undirected exploration such as maximum-entropy objectives.
In this paper we investigate the missing synergy between off-policy continuous control and directed exploration.

We find that BBE is poorly suited to the few-sample regime due to slowly-decaying bias in the learned policy and slow adaptation to the non-stationary exploration bonus.
Bias due to optimizing a reward function other than the task reward leads a policy trained with BBE to exhibit poor performance until the bonus decays toward zero.
Meanwhile, the non-stationary (continually decreasing) exploration bonus cannot necessarily be optimized by a fixed policy, violating one of the core assumptions of RL.
This leads to slow exploration as the policy adapts only gradually, especially in the off-policy case where replay buffers will contain stale rewards.
These observations underline research by \citet{Taiga2020On} showing that across the ALE, no BBE algorithm outperforms undirected $\varepsilon$-greedy exploration.

We demonstrate that bias and slow coverage are the culprits of BBE's lackluster performance by proposing a new exploration algorithm, \algname{} (\algshort{}), which addresses these limitations.
\algshort{} decouples the learning of a \emph{task policy}, which is trained to maximize the true task reward, and an \emph{exploration policy}, which maximizes only the exploration bonus.
Both policies are trained off-policy using data collected according to the product of the two policy distributions.
Unlike the policy learned by BBE, \algshort{}'s task policy is always unbiased in the sense that it reflects the current belief about the optimal action in each state.
Furthermore, this decoupling allows \algshort{} to aggressively update the exploration policy without affecting the convergence of the task policy, thereby adapting more rapidly to the changing exploration bonus.

We perform experiments using policies based on Q-learning \citep{Sutton2018ReinforcementLA,mnih2015human} on toy tasks and soft actor-critic (SAC) \citep{haarnoja2018softB} on larger-scale tasks from the DeepMind Control Suite \citep{tassa2018deepmind}.
Our results show that on tasks with dense rewards and uniform resets, BBE often performs worse than the underlying policy-learning algorithm while \algshort{} incurs no cost for exploring.
On tasks with more natural resets and sparse rewards, \algshort{} covers the state space more rapidly than BBE and reaches peak performance in a fraction of the samples required with undirected exploration.
In total, \algshort{} strictly outperforms undirected exploration while solving many sparse environments just as fast as dense ones.

\section{Background}

\subsection{Notation}

A Markov decision process (MDP) $\mdp$ consists of a tuple $(\mathcal{S}, \mathcal{A}, P, R, \gamma)$, where $\mathcal{S}$ is the state space, $\mathcal{A}$ is the action space, $P$ is the transition function mapping $\mathcal{S} \times \mathcal{A}$ to distributions on $\mathcal{S}$, $R$ is the scalar reward function on $\mathcal{S} \times \mathcal{A}$, and $\gamma$ is the discount factor.
We use lower-case $(s, a, r)$ to refer to concrete realizations of states, actions, and rewards.
We use $\mdp_f$ to denote the MDP $\mdp$ with the original reward function $R$ replaced by another function $f$.
For convenience we assume exploration rewards are within $[0, 1]$, and we define $\bar{r} = \nicefrac{1}{1 - \gamma}$, which is the maximum discounted value possible.

\subsection{Bonus-based exploration: a recipe for exploration in deep RL} \label{sec:intrinsic_rewards}

Bonus-based exploration has emerged as the standard framework for exploration in the deep reinforcement learning community.
In this framework, an agent learns in a sequence of MDPs $\widetilde{\mdp} = \{\mdp_{\widetilde{R}_n} \}_{n=1}^N$ where the reward function $\widetilde{R}_n$ changes as a function of each transition.
A typical choice is $\widetilde{R}_n = R + R^+_n$, where $R^+_n$ is an exploration bonus which measures the ``novelty'' of a transition $(s, a, s')$ given the history of all transitions up to $n$.
After taking each transition $(s, a, s')$, the reward $\widetilde{r} = \widetilde{R}_n(s, a, s')$ is calculated and the tuple $(s, a, s', \widetilde{r})$ is added to a replay dataset $D$.
The agent optimizes its reward in this (non-stationary) MDP $\widetilde{\mdp}$ via some model-free RL algorithm operating on the replay dataset.
The realization of a particular algorithm in this family amounts to defining a novelty function and picking a model-free RL algorithm \citep{Stadie2015IncentivizingEI,houthooft2016vime,bellemare2016unifying,pathak2017curiosity,Tang2017Exploration,burda2018exploration,Machado2020CountBasedEW}.
We illustrate this recipe in \cref{alg:algorithm1}.

\paragraph{Pseudo-counts.}
Building upon theoretically-motivated exploration methods for discrete environments \citep{strehl2008analysis}, \citet{bellemare2016unifying} proposed to give exploration bonuses based on a \emph{pseudo-count} function $\hat N$.
A pseudo-count has two key properties.
Like a true count, a pseudo-count increases by 1 each time a state (or state-action pair) is visited.
Unlike a true count, a pseudo-count generalizes across states and actions; that is, when a state $s$ is visited, the pseudo-count for nearby states $s + \varepsilon$ may increase as well.

\begin{figure*}
\begin{minipage}[b]{0.48\textwidth}
\begin{algorithm}[H]
    \small
    \centering
    \caption{Bonus-based exploration}\label{alg:algorithm1}
    \begin{algorithmic}[1] 
        \Require{replay dataset $D$, policy $\pi$, bonus $R^+_n$}
        \item[]
        \State $n \gets 0$
        \Repeat
            \For{one episode}
                \item[]
                \item[]
                \State Collect $(s, a, s', r) \sim P(s, \pi(s))$
                \State $\widetilde{r} \gets r + R^+_n(s, a, s')$
                \State $D \gets D \cup (s, a, s', \widetilde{r})$
                \State $R^+_{n+1} \gets \text{Update}(R^+_n, (s, a, s'))$
                \State $n \gets n + 1$
            \EndFor
            \State Train $\pi$ with samples from $D$
        \Until{$n = N$}
    \end{algorithmic}
\end{algorithm}
\end{minipage}
\hfill
\begin{minipage}[b]{0.48\textwidth}
\begin{algorithm}[H]
    \small
    \centering
    \caption{\algshort{}}\label{alg:algorithm2}
    \begin{algorithmic}[1]
        \Require{replay dataset $D$, temperature $\tau$,}
        task policy $\pitask$, exploration policy $\piexplore$, bonus $R^+_n$
        \State $n \gets 0$
        \Repeat
        \For{one episode}
            \State Update $\piexplore$ on $\mdp_{R^+_n}$
                \State Set $\beta(a|s) \propto \pitask(a|s) \cdot \piexplore(a|s)$ 
                \State Collect $(s, a, s', r) \sim P(s, \beta(s))$
                \item[]
                \State $D \gets D \cup (s, a, s', r)$
                \State $R^+_{n+1} \gets \text{Update}(R^+_n, (s, a, s'))$
                \State $n \gets n + 1$
            \EndFor
            \State Train $\pitask$ with samples from $D$
        \Until{$n = N$}
    \end{algorithmic}
\end{algorithm}
\end{minipage}
\caption{
Comparison of classic bonus-based exploration (BBE) with our method (\algshort{}).
BBE computes exploration bonuses at the time of visiting a transition, adds them to the real rewards, and uses a replay buffer of experience to learn a policy.
\algshort{} separates the exploration policy $\piexplore$ from the task policy $\pitask$, allowing $\pitask$ to be an unbiased estimate of the optimal policy throughout training.
It always uses the \emph{current} exploration reward function $R_n^+$ when updating the exploration value function, and is fast-adapting to deal with the non-stationary bonus MDP.
}
\end{figure*}

\section{Limitations of bonus-based exploration}

\begin{figure}[h]
    \centering
    \begin{subfigure}[b]{0.49\textwidth}
        \centering
        \includegraphics[width=0.8\textwidth]{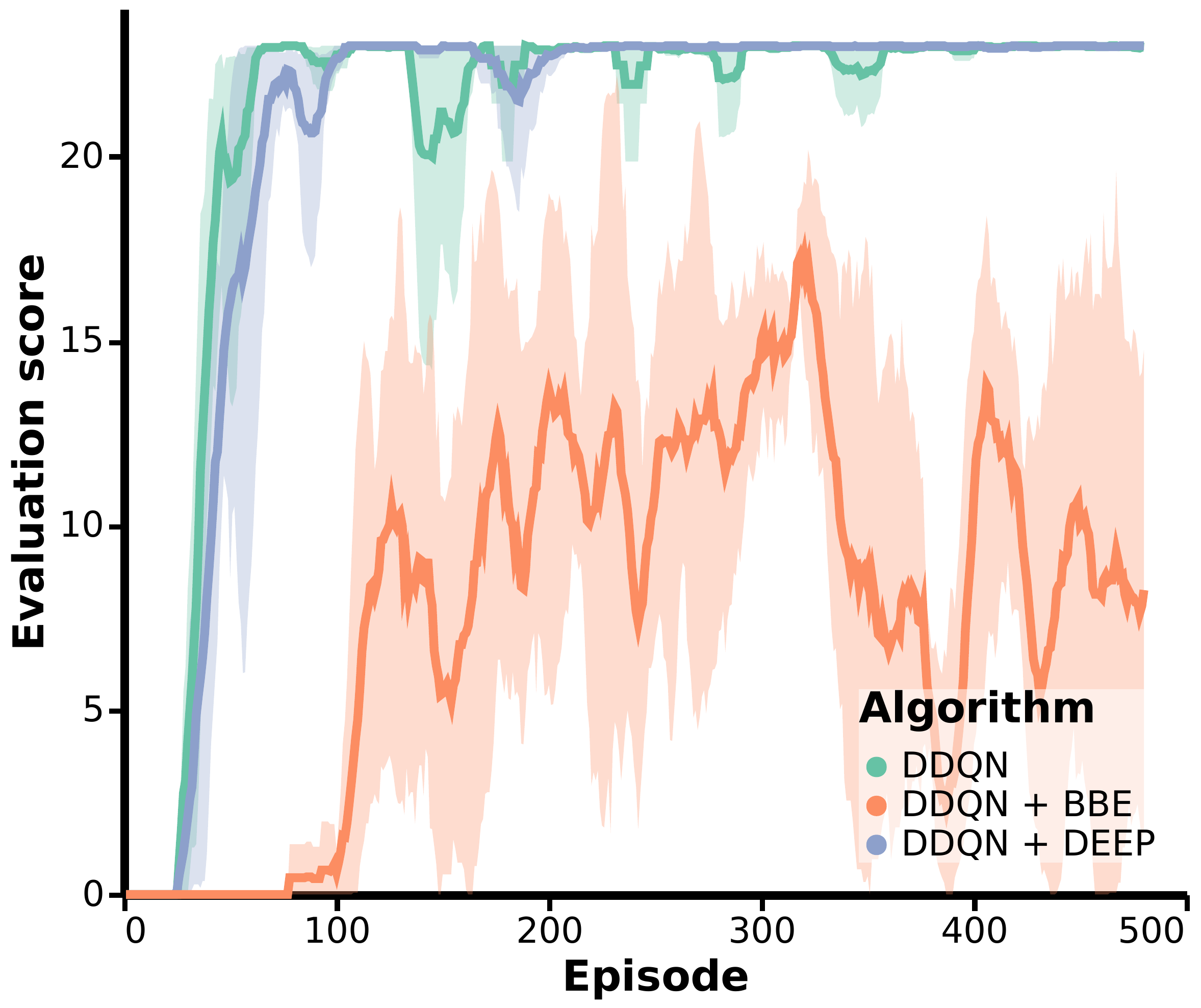}
        \caption{Reward}
    \end{subfigure}
    \hfill
    \begin{subfigure}[b]{0.49\textwidth}
        \centering
        \includegraphics[width=0.8\textwidth]{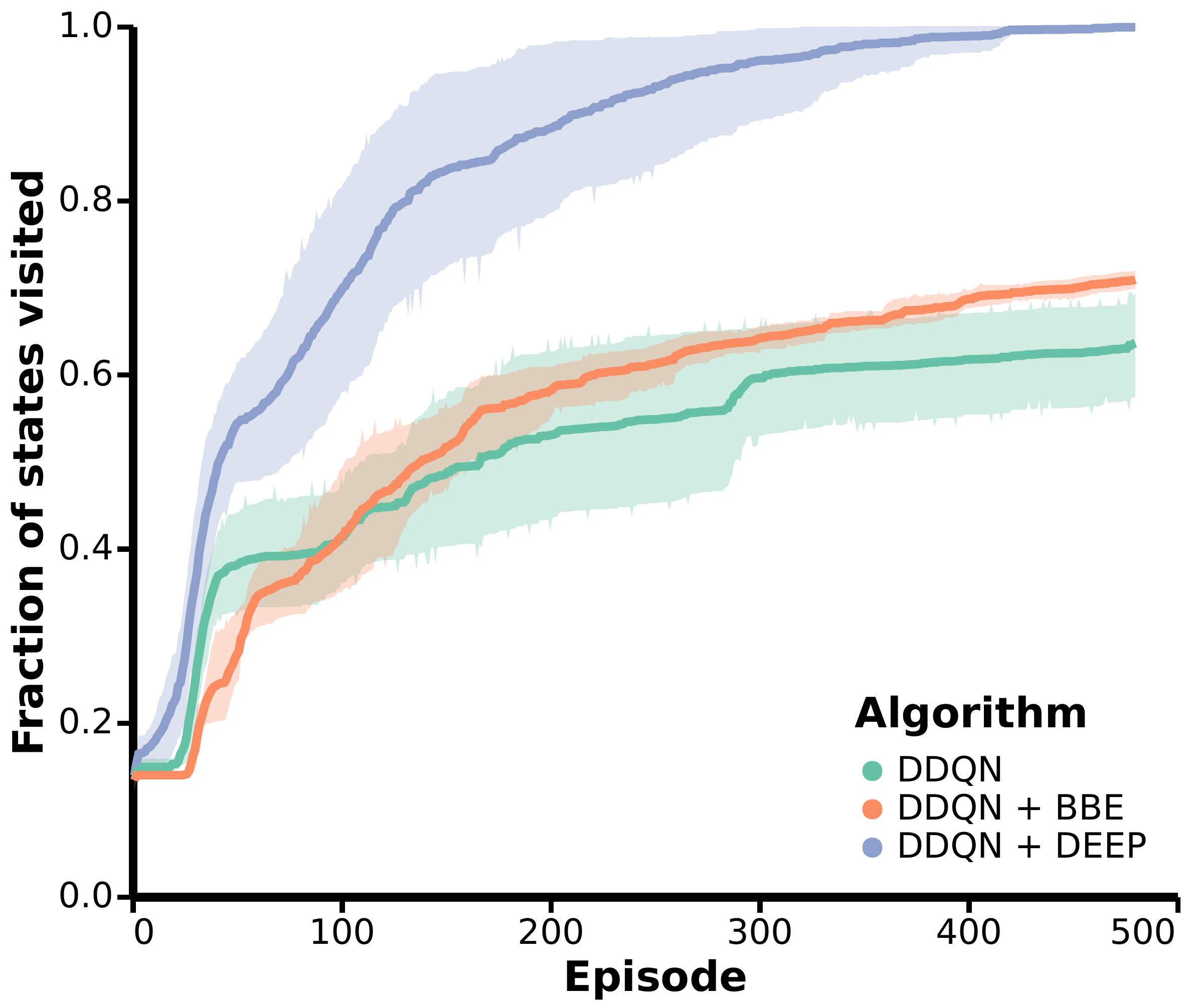}
        \caption{Coverage}
    \end{subfigure}
    \caption{Experiments in a 40x40 grid-world environment with one goal state, where learning algorithms were warm-started with 20 episodes of data from a skilled policy.
    \textbf{(a)} With enough signal to find the goal, DDQN (Double DQN, \citet{Hasselt2016DeepRL}) alone rapidly converges to the optimal policy. BBE introduces bias, causing the policy to continually explore. Our method, \algshort{}, learns the task policy just as rapidly as DDQN alone.
    \textbf{(b)} Though it performs well, DDQN simply goes to the goal during each train episode and does not explore other options. BBE continues to seek out new states at a slow but steady rate. \algshort{} explores far more than BBE during data collection despite simultaneously performing just as well as DDQN at evaluation time.}
    \label{fig:gridworld_warmstart}
\end{figure}

The bonus-based exploration algorithm, illustrated in \Cref{alg:algorithm1}, has two weaknesses which limit its usefulness for sample-efficient policy learning.

\paragraph{Bias with finite samples.}
Because they estimate the optimal policy on the modified MDP $\mdp'$, bonus-based exploration algorithms learn biased policies as long as the exploration bonus is nonzero.
According to theory, the exploration bonus should be scaled larger than is done in practice \citep{strehl2008analysis} and decay slower than $\nicefrac{1}{N(s)}$ \citep{Kolter2009NearBayesianEI} in order to guarantee convergence to the optimal policy.
This behavior, shown in \Cref{fig:gridworld_warmstart}, can result in slow convergence to the optimal policy and substantially biased policies after a practically feasible number of samples.

\begin{wrapfigure}{r}{0.4\textwidth}
    \begin{center}
        \includegraphics[width=0.35\textwidth]{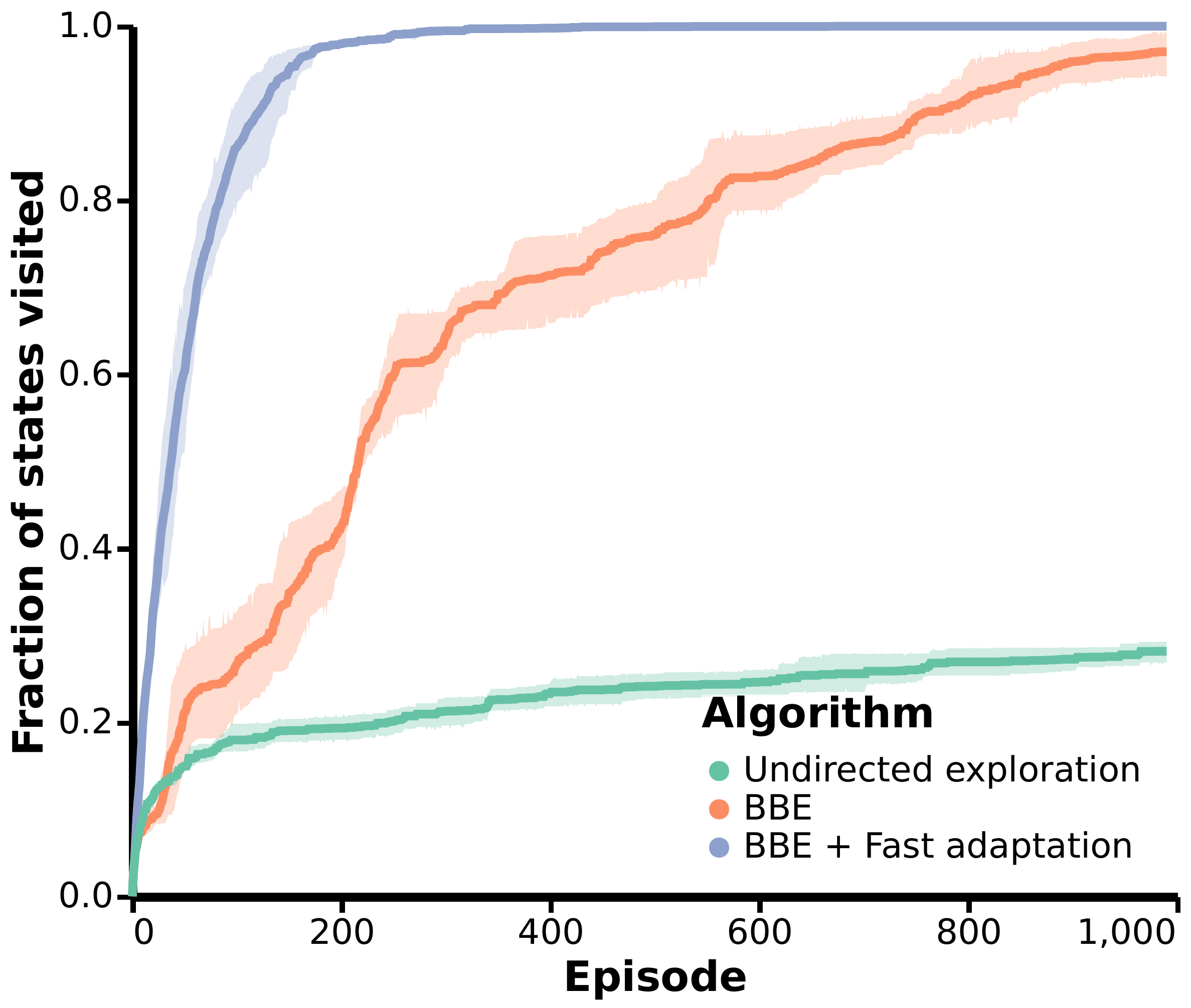}
    \end{center}
    \caption{Pure exploration}
    \label{fig:gridworld_visits}
\end{wrapfigure}


\paragraph{Slow adaptation to changing rewards.}
Algorithms in this family update the policy according to the schedule of the underlying model-free RL algorithm -- for example at the end of each episode.
This works well for the stationary MDPs that these algorithms were developed for, but the modified MDP $\mdp'$ which represents the exploration problem is non-stationary.
This leads to an agent which determines the most novel state and then stays there for an entire episode.
This degenerate behavior leads to potentially exploring only a single state per episode instead of visiting a sequence of new states as the reward function evolves.\footnote{Some implementations of bonus-based exploration may update the policy within an episode, for example via a single gradient step per environment step on transitions sampled i.i.d.~from a replay.
However, such a small update is typically not enough to change the qualitative behavior of the agent and adapting to the changing MDP has not been an emphasis in prior work.}
The use of replay buffers compounds this effect, since algorithms in this family compute exploration rewards at the time the transition is collected, rather than when it is used.
An algorithm which is unaware of the non-stationary nature of the MDP will maximize the return on this mixture of reward functions rather than the reward that incorporates the current bonus, resulting in slow coverage of the environment.
\Cref{fig:gridworld_visits} shows uniform random actions, BBE, and BBE with the fast adaptation scheme we propose in \Cref{sec:fast_adaptation} all exploring in a 40x40 grid-world without rewards.
While BBE covers the state space much faster than undirected exploration, it is unnecessarily slow.
See Appendix \ref{sec:gridworld-vis} for visualizations.

\section{Decoupled exploration for sample-efficient control}
In this section, we describe a new algorithm called \algname{} (\algshort{}) which addresses the limitations of BBE.
The core insight is that by leveraging off-policy RL algorithms, \algshort{} can learn two policies from the same replay: a task policy $\pitask$, which maximizes the reward on the original MDP $\mdp_R$, and an exploration policy $\piexplore$, which maximizes only the reward on the bonus MDP $\mdp_{R^+_n}$.
This decoupling serves two purposes.
First, it enables good performance even before exploration is complete by using $\pitask$ at test time.
Second, it allows $\piexplore$ to be updated aggressively in order to more closely match the non-stationary bonus MDP; unlike $\pitask$, it is not important that $\piexplore$ converge exactly to an optimal policy.

Like BBE, \algshort{} is a family of algorithms related by their structure; a particular algorithm in this family consists of a choice of an exploration reward function and an off-policy RL algorithm for learning each policy.
Throughout this work, we use a pseudo-count based exploration reward.
For discrete tasks we use Double DQN (DDQN, \citet{Hasselt2016DeepRL}) and Boltzmann policies.
For tasks with continuous actions we use soft actor-critic (SAC) for $\pitask$ and a DDQN policy for $\piexplore$.

\subsection{Pseudo-count estimation} \label{sec:kernel_counts}
Following \citet{bellemare2016unifying}, we use an exploration reward derived from pseudo-counts.
Instead of the high-dimensional pixel observations of the ALE, Control Suite has low-dimensional ($<$100-D) observations corresponding to joint and object locations and velocities.
This lower dimensionality renders extracting pseudo-counts from a density estimator unnecessary, and in our experiments we specify a pseudo-count function based on kernels.
For a real-valued state-action pair $x = [s, a]$ and a set of previous observations $\{x_i\}_{i=1}^n$, define the pseudo-count of $x$ and the exploration reward as
\begin{align}
    \hat N_n(x) = \sum_{i=1}^n k(x_i, x) && R^+_n(s, a) = \hat N_n \big([s, a] \big)^{-1/2}
\end{align}
where $k$ is a kernel function scaled to have a global maximum $k(x, x) = 1$.
This satisfies the key requirement for a pseudo-count function, namely that a visit to a state $x$ increases $\hat N(x)$ by 1 and $\hat N(x')$ by at most 1 for any other state $x'$.
In our experiments we use a Gaussian kernel (scaled to have a maximum value of 1) with diagonal covariance.
For implementation details see Appendix \ref{sec:count_implementation}.

\subsection{Separating task and exploration policies}

\algshort{} uses two separate policies.
Each is trained off-policy using transitions sampled from the replay buffer; $\pitask$ is updated using the rewards from $R$ logged in the replay, while $\piexplore$ is updated using rewards from $R^+_n$.
Since $\pitask$ is trained only on the rewards for the true task, it is unbiased in the sense that it reflects the current best estimate of the optimal policy.
This stands in contrast to BBE policies, which optimize the sum of task and exploration rewards and thus represent a biased estimate of the optimal task policy until the exploration rewards go to zero.
Our method is agnostic to the choice of algorithm and policy parameterization.
However, it will be most effective with policy learning algorithms that work well when trained far off-policy and produce high-entropy policies (e.g. policies which cover all optimal actions).
For these reasons we use the state-of-the-art maximum-entropy algorithm SAC to learn $\pitask$ in environments with continuous actions. For experiments with discrete action spaces we forego explicitly learning the task policy $\pitask$; instead we learn the task Q-function via DDQN \citep{Hasselt2016DeepRL} and define the task policy as $\pitask(a | s; Q) \propto \exp(Q(s, a) / \tau)$, where $\tau > 0$ becomes a hyperparameter -- we refer to the appendix material for details.

\subsection{Fast-adapting exploration policy} \label{sec:fast_adaptation}
The non-stationary nature of the exploration reward function poses a challenge to typical model-free RL algorithms, which assume a fixed reward function.
BBE methods update a single policy using a replay buffer which, at a step $n$, contains rewards from a mixture of bonus reward functions $\{R_1^+, \ldots, R_n^+\}$, computed using different past novelty or count estimates.\footnote{See e.g. the code from \citet{Machado2020CountBasedEW}:  \url{https://github.com/mcmachado/count_based_exploration_sr/blob/master/function_approximation/exp_eig_sr/train.py\#L204}}
This results in slow adaptation to the non-stationary objective of exploration.
\algshort{} makes two changes to mitigate the impact of the non-stationarity in the exploration reward function.

First, \algshort{} leverages access to $R^+_n$ to compute exploration rewards when they are needed to update $\piexplore$ rather than when the transition is collected.
We choose to use Q-learning rather than a more sophisticated algorithm in order to learn $\piexplore$ as rapidly as possible; with changing rewards, using a policy to amortize the maximization of a value function as in SAC or DDPG would slow down learning.
We represent $\piexplore$ directly as a Boltzmann policy of this exploration Q-function $\qex$:
\begin{align}
    \piexplore(a \mid s) \propto \exp \left\{ \frac{\qex(s, a)}{\tex} \right\},
\end{align}
where $\tex$ is a temperature hyperparameter.

Second, by decoupling $\piexplore$ from $\pitask$, \algshort{} unlocks the ability to update $\piexplore$ more aggressively without affecting $\pitask$'s convergence to the optimal policy.
This enables the exploration policy to more rapidly adapt to the non-stationary exploration reward.
In our experiments we achieve this by using a larger learning rate and more updates per environment step than is usually done; future work might investigate more sophisticated schemes such as prioritized sweeping \citep{Moore1993PrioritizedSR} or prioritized experience replay \citep{Schaul2016PrioritizedER}.
To improve stability we use DDQN updates and clip Q targets at $\bar{r}$, the maximum discounted exploration value.

\paragraph{Optimism.} \label{sec:optimistic}
Further adapting $\piexplore$ to the unique properties of the exploration reward function, we propose to leverage optimism in its updates and actions.
We propose to make $\qex$ optimistic by leveraging the pseudo-count function in a manner similar to that proposed by \citet{Rashid2020OptimisticEE}.
We assume that the value function is trustworthy for transitions with very large counts, and very untrustworthy for transitions with near-zero counts.
When the count is zero we impose an optimistic prior which assumes the transition will lead to a whole episode of novel transitions; as the count increases we interpolate between this prior and the learned value function using a weighting function:
\begin{align} \label{eq:optimism}
\qex^+(s, a) = w(s, a) \cdot \qex(s, a) + (1 - w(s, a))  \cdot \bar{r}, \qquad w(s, a) = \frac{\sqrt{N(s, a)}}{\sqrt{N(s, a) + c}}
\end{align}
where $\bar{r} = \nicefrac{1}{1-\gamma}$, is the maximum discounted return in the bonus MDP and $c$ is a small constant representing how many counts' worth of confidence to ascribe to the optimistic prior.
We use this optimistic $\qex^+$ for computing targets for Bellman updates and for computing $\piexplore$ (Eq. \ref{eq:behavior_policy}).
For details of the implementation of the fast-adapting $\piexplore$, see Appendix \ref{sec:fast_updates_appendix}.

\subsection{Product distribution behavior policy}
A good behavior policy should attempt to explore all of the transitions which are relevant for learning the optimal policy.
This entails a trade-off between taking actions which are more novel and ones which are more likely to be relevant to a high-performing policy.
\algshort{} encodes this by representing the behavior policy as a product of the task policy $\pitask$ and the pure-exploration policy $\piexplore$:
\begin{align} \label{eq:behavior_policy}
\beta(a \mid s) \propto \pitask(a \mid s) ~ \piexplore(a \mid s).
\end{align}

The choice to parameterize $\beta$ as a factored policy was made for its simplicity and ease of off-policy learning.
Alternative formulations for making this trade-off while preserving the unbiased task policy are possible, and we view the form of our proposed behavior policy as just one option among many.
One alternative would be interleaving the behavior of multiple policies within one episode, akin to e.g. Scheduled Auxiliary Control \citep{Riedmiller2018LearningBP}.

In order to approximately sample from this behavior policy, we use self-normalized importance sampling with $\pitask$ as the proposal distribution:
\begin{enumerate}
    \item Draw $k$ samples $a_1, \ldots, a_k$ from $\pitask$
    \item Evaluate $\piexplore(a_i \mid s)$ for each $i \in 1 \ldots k$
    \item Draw a sample from the discrete distribution $p(a_i) = \piexplore(a_i \mid s) / \sum_{i'} \piexplore(a_{i'} \mid s)$.
\end{enumerate}
Note that since the proposal distribution is $\pitask$, the $\pitask$ terms in computing weights cancel and only the $\piexplore$ terms remain.
Importance weighting is consistent in the limit of $k \to \infty$ but introduces bias towards $\pitask$ \citep{Vehtari2015ParetoSI}.
With small $k$, this bias makes it unlikely that $\beta$ will select actions that are very unlikely under $\pitask$; roughly speaking, this procedure selects the ``most exploratory'' action in the support of the task policy.
This may act as a backstop to prevent the behavior from going too far outside the task policy to be useful.
\algshort{} works best when $\pitask$ is trained in a way that preserves variance in the policy (e.g. SAC's target entropy), enabling the behavior policy to select exploratory actions. In discrete action spaces we additionally use self-normalized importance sampling -- using a uniform proposal over actions -- to obtain samples from $\pitask$ in step 1.


\section{Experiments}

In this section we perform experiments to give insight into the behavior of undirected exploration, BBE, and \algshort{}.
First we perform a set of investigative experiments to probe how \algshort{} interacts with environments with different reward structures.
Then we perform experiments on pairs of benchmark continuous control tasks with easy and hard exploration requirements.

\begin{figure}[h]
    \centering
    \begin{subfigure}[b]{0.49\textwidth}
        \centering
        \includegraphics[width=0.8\textwidth]{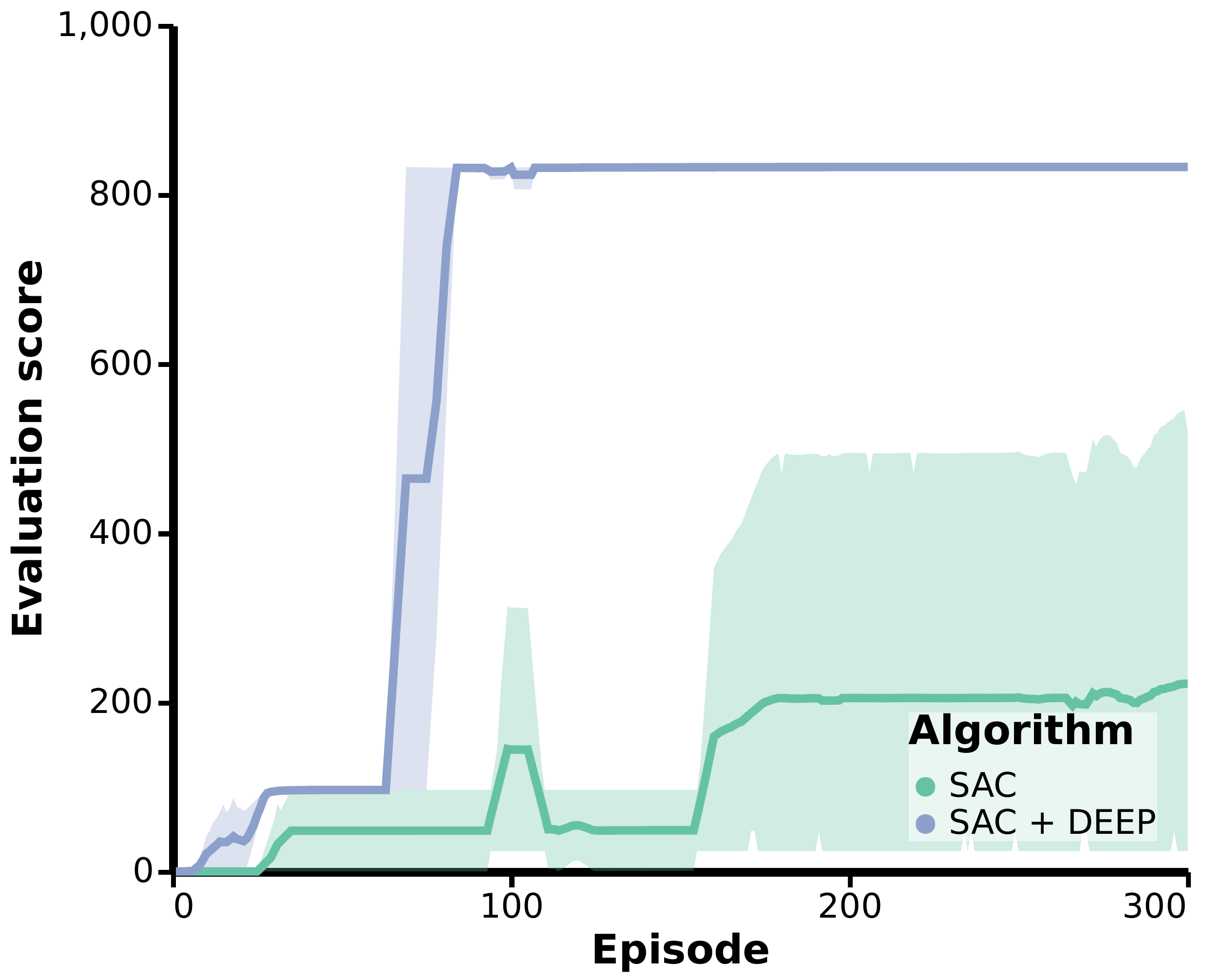}
        \caption{Local optimum environment} \label{fig:hallway_local_optimum}
    \end{subfigure}
    \hfill
    \begin{subfigure}[b]{0.49\textwidth}
        \centering
        \includegraphics[width=0.8\textwidth]{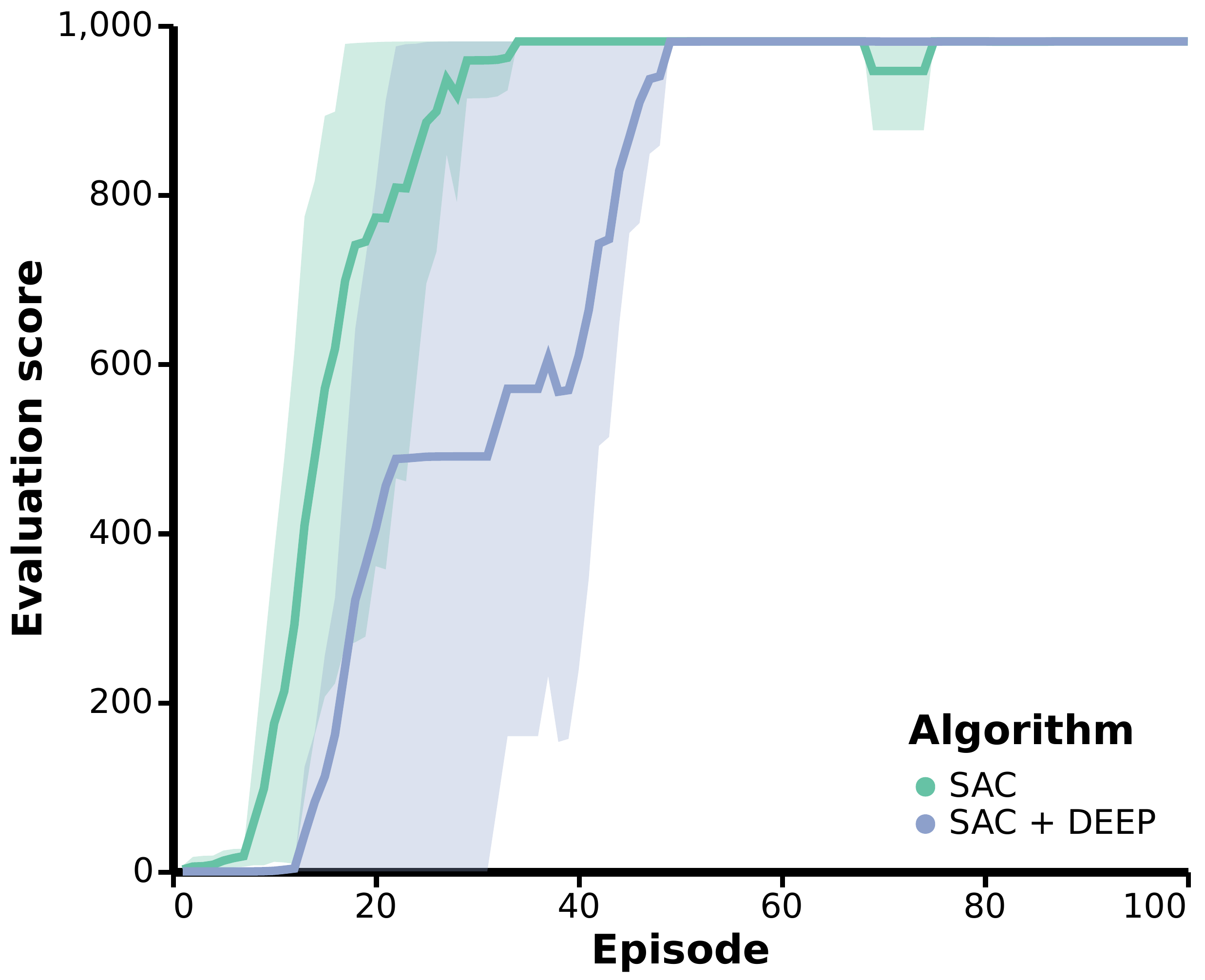}
        \caption{Adversarial environment}\label{fig:inverse_distractor}
    \end{subfigure}
    \caption{Two environments illustrating different reward structures.
    \textbf{(a)} An environment with a locally-optimal goal (reward 0.1) near the start state. SAC finds this nearby goal, but doesn't explore far enough to find the real goal (reward 1.0). When trained with \algshort{}, it finds the distractor goal but moves on to the real goal.
    \textbf{(b)} An adversarial environment for \algshort{} which has a very small goal state close to the start state, making it easy to find with random actions but hard with directed exploration.}
    \label{fig:hallway}
\end{figure}

\subsection{Investigative experiments}

We construct a simple MuJoCo \citep{todorov2012mujoco} environment called Hallway to look more closely at how exploration interacts with reward structure.
This environment consists of a long narrow 2D room with the agent controlling the velocity of a small sphere, which starts each episode at one end of this hallway.
The following two experiments share dynamics and differ only in their rewards.

\paragraph{Local optima.}
A valuable role for exploration is enabling an agent to escape from locally optimal behavior.
To test this, we add two goal states with shaped rewards to the Hallway environment.
The first is close to the start state but only provides reward at most 0.1, while the second is at the far end of the hallway but provides reward 1.0.
\Cref{fig:hallway_local_optimum} shows that exploration using \algshort{} allows the agent to quickly find its way to the faraway optimal reward while SAC gets stuck in the local optimum.

\paragraph{Limitations.}
\algshort{} covers states quickly, but there is no such thing as a universally optimal exploration strategy.
For example, there exist environments for which the random walk dynamics of undirected exploration find the optimal strategy faster than uniform state coverage.
\Cref{fig:inverse_distractor} provides one such example: a Hallway environment with a very small goal state close to the start state.
SAC discovers this goal faster than SAC + \algshort{}, though \algshort{} does eventually find it as well.

\subsection{Benchmark experiments}
Next we provide experiments based on DeepMind Control Suite \citep{tassa2018deepmind}, a standard benchmark for continuous control RL algorithms.
We introduce versions of several environments which are modified to remove the accommodations that make them solvable without exploration, then provide results of SAC, SAC + BBE, and SAC + \algshort{} on the original and modified environments.

\subsubsection{Environments for evaluating exploration}
While Control Suite has driven great progress in policy learning, it was not designed to evaluate an agent's exploration capabilities; in fact, the included environments were selected to be solvable by algorithms with only undirected exploration. From that work:
\begin{quote}
    We ran variety of learning agents (e.g. Lillicrap et al. 2015; Mnih et al. 2016) against all tasks, and iterated on each task’s design until we were satisﬁed that
    [\ldots]
    the task is solved correctly by at least one agent. \citep{tassa2018deepmind}
\end{quote}
Control Suite avoids the need for directed exploration via two mechanisms. 
First, in many environments the start state distribution is sufficiently wide (e.g. uniform over reachable states) to guarantee that any policy will see high-value states.\footnote{Some environments, such as Manipulator, additionally start a fraction of episodes at the goal state.}
Second, some environments have rewards shaped to guide the agent towards better performance (e.g. a linearly increasing reward for forward walking speed).

To construct a benchmark for continuous control with exploration, we selected four environments with different objectives (manipulation and locomotion, single-objective or goal-conditional) and observation dimensions (6-24).
We then created ``exploration'' versions of these environments with restricted start state distributions and sparse rewards.
The original environments and their exploration versions together form a benchmark which measures an algorithm's exploration ability and policy convergence.
Environment details are in Appendix \ref{sec:environments_appendix}.

\subsubsection{Algorithms}
We include experiments on these eight benchmark tasks with three algorithms: SAC \citep{haarnoja2018softB} with no additional exploration; BBE with SAC for the policy learner; and \algshort{} with SAC for $\pitask$ and DDQN for $\piexplore$.
BBE and \algshort{} use the pseudo-count reward described in \Cref{sec:kernel_counts} and the SAC implementation is that of \citet{pytorch_sac} with no hyperparameter changes.

The kernel-based exploration bonus used for BBE and \algshort{} requires a scaling law to set the kernel variance as a function of the observation dimension.
We adapt the scaling relationship from \citep{Henderson2012NormalRB} (see Appendix \ref{sec:count_implementation}).
BBE has an additional hyperparameter for the scale of the bonus.
We performed a sweep with values in $\{ 10^{-2}, 10^{-1}, 1, 10 \}$ and found that $1$ performed best overall.
This setting, which makes the maximum bonus equal to the maximum environment reward, ensures that visiting a new state remains the best option until the true goal state is discovered.
Further implementation details are available in Appendix \ref{sec:appendix_benchmark_implementation}.
We additionally performed an ablation which, like \algshort{}, learns separate Q functions for the two rewards, but which learns one policy to maximizes the sum of their Q values.
In our experiments (available in Appendix \ref{sec:benchmark_results_appendix}) this ablation never outperforms BBE, so for clarity we exclude it here.

\subsubsection{Results}
\begin{figure}[t]
    \centering
    \begin{subfigure}[b]{0.24\textwidth}
        \centering
        \includegraphics[width=\textwidth]{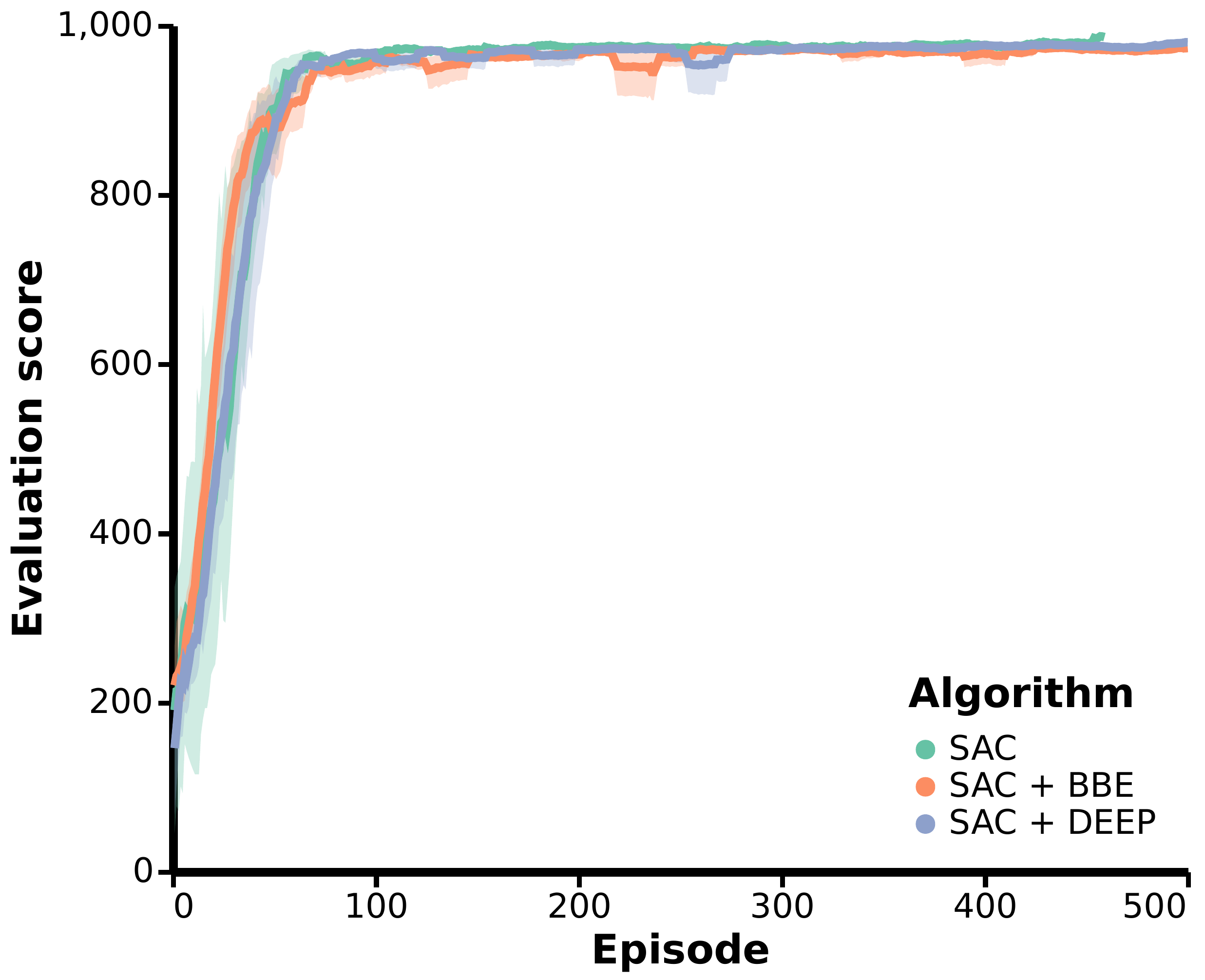}
        \caption{Ball-in-cup}
    \end{subfigure}
    \begin{subfigure}[b]{0.24\textwidth}
        \centering
        \includegraphics[width=\textwidth]{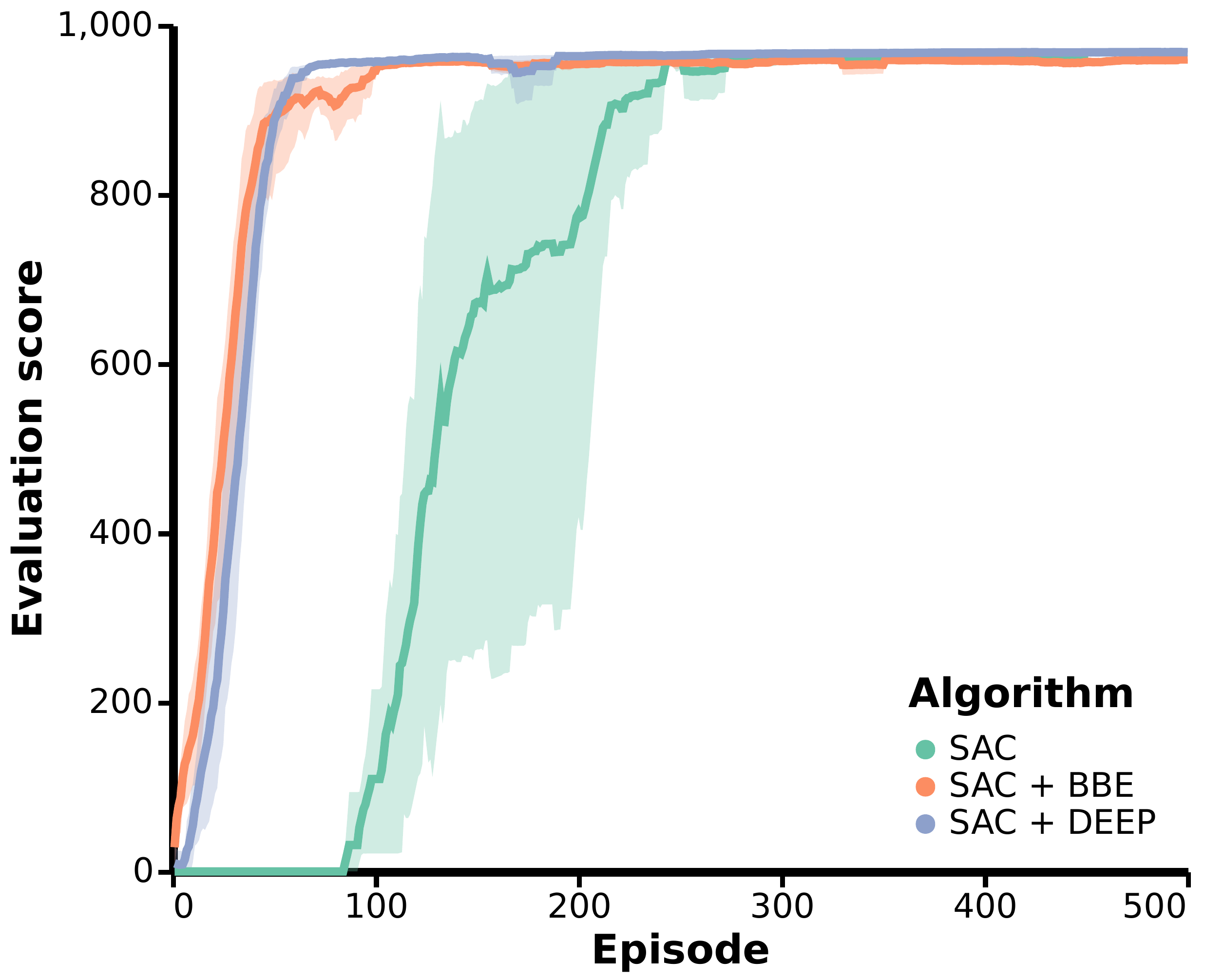}
        \caption{Ball-in-cup explore}
    \end{subfigure}
    \hfill
    \begin{subfigure}[b]{0.24\textwidth}
        \centering
        \includegraphics[width=\textwidth]{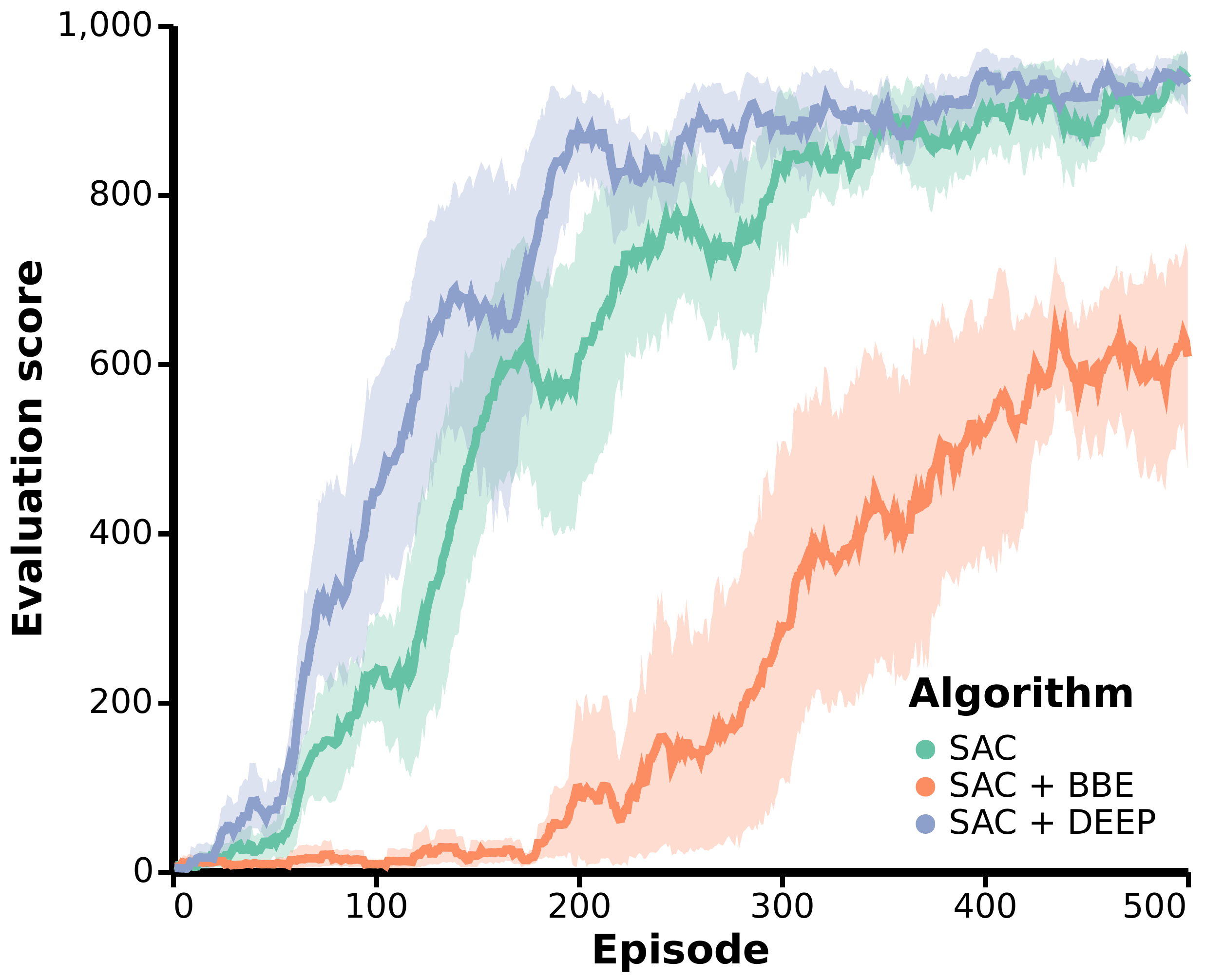}
        \caption{Reacher}
    \end{subfigure}
    \begin{subfigure}[b]{0.24\textwidth}
        \centering
        \includegraphics[width=\textwidth]{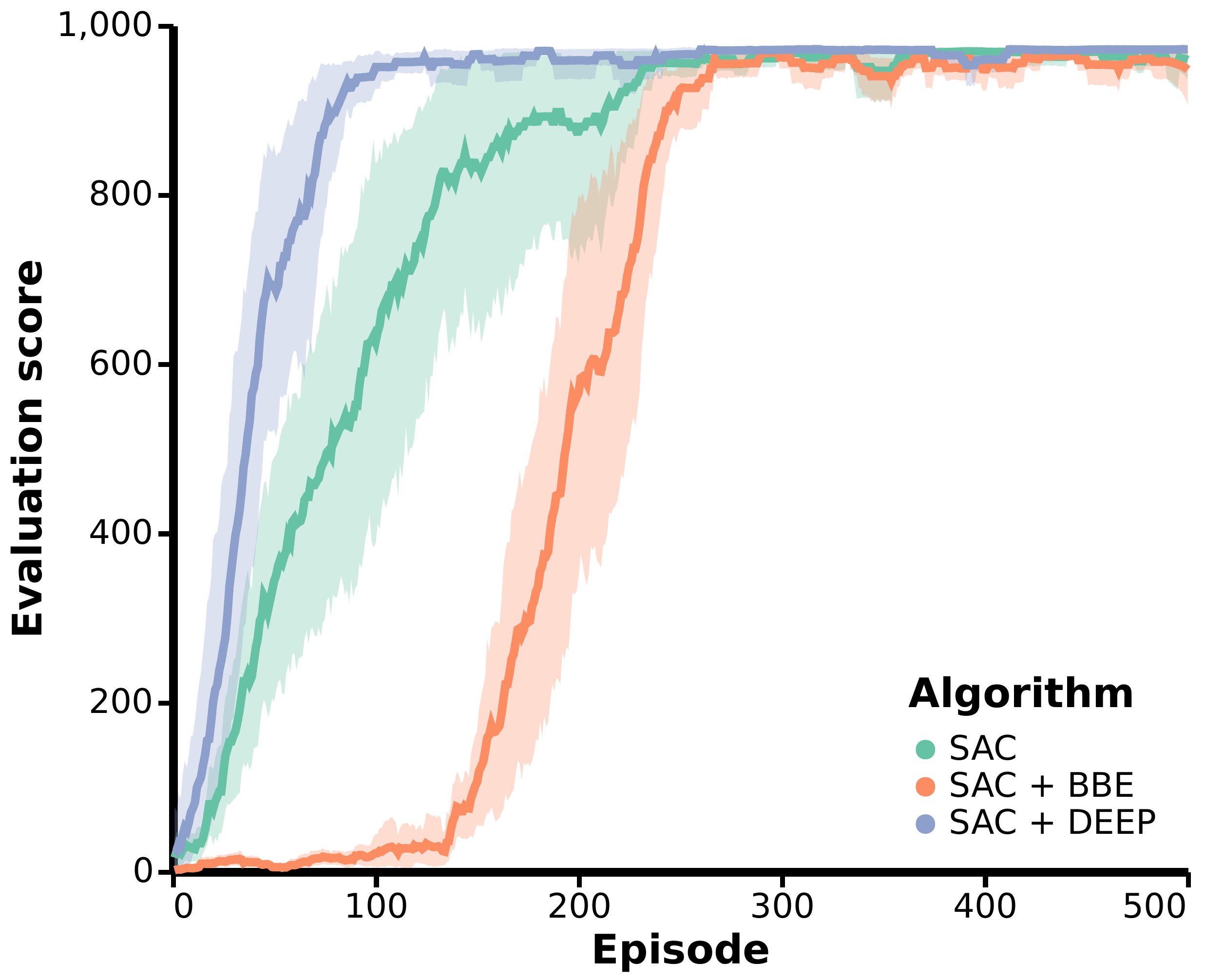}
        \caption{Reacher explore}
    \end{subfigure}
    \vspace{1em}

    \begin{subfigure}[b]{0.24\textwidth}
        \centering
        \includegraphics[width=\textwidth]{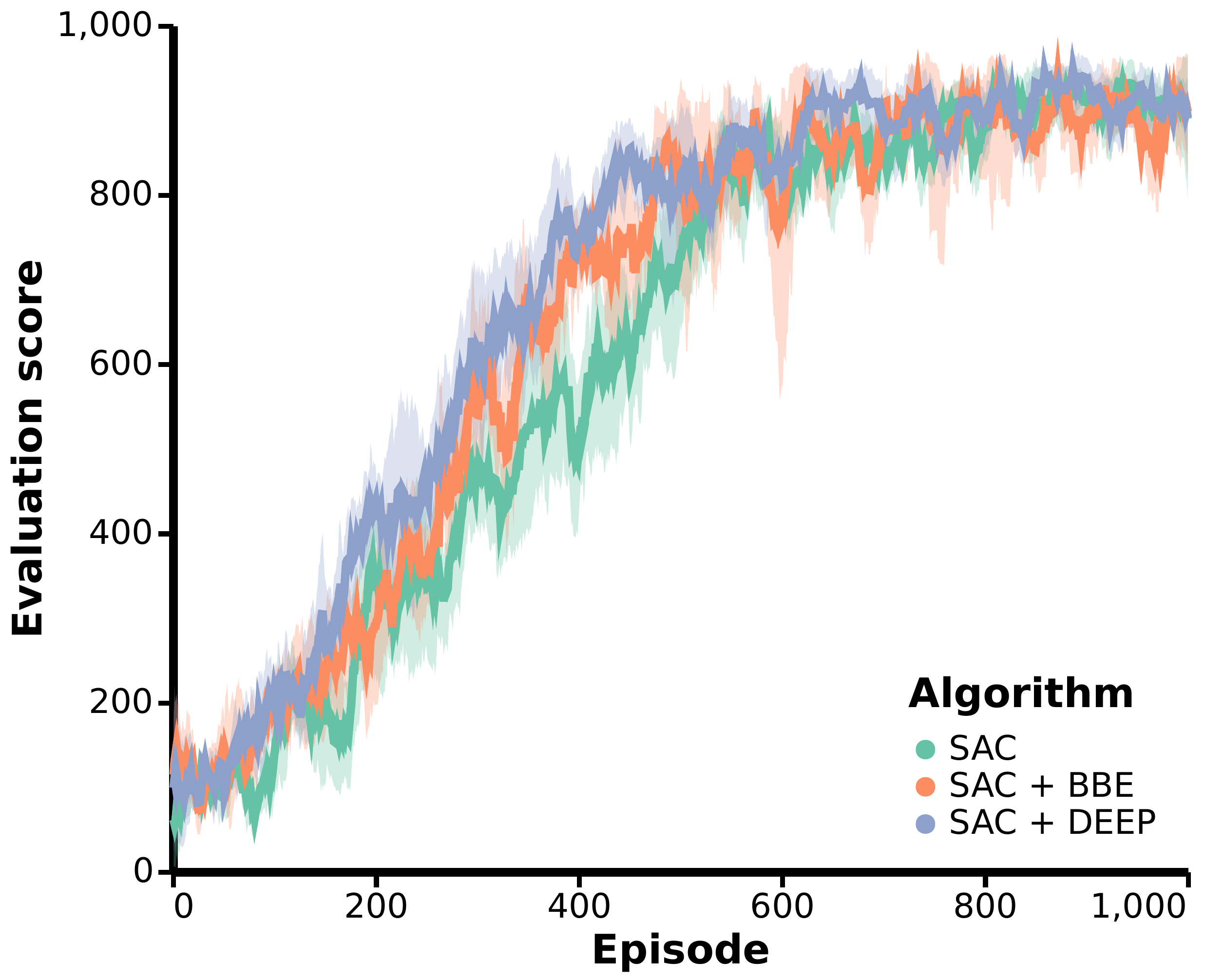}
        \caption{Finger}
    \end{subfigure}
    \begin{subfigure}[b]{0.24\textwidth}
        \centering
        \includegraphics[width=\textwidth]{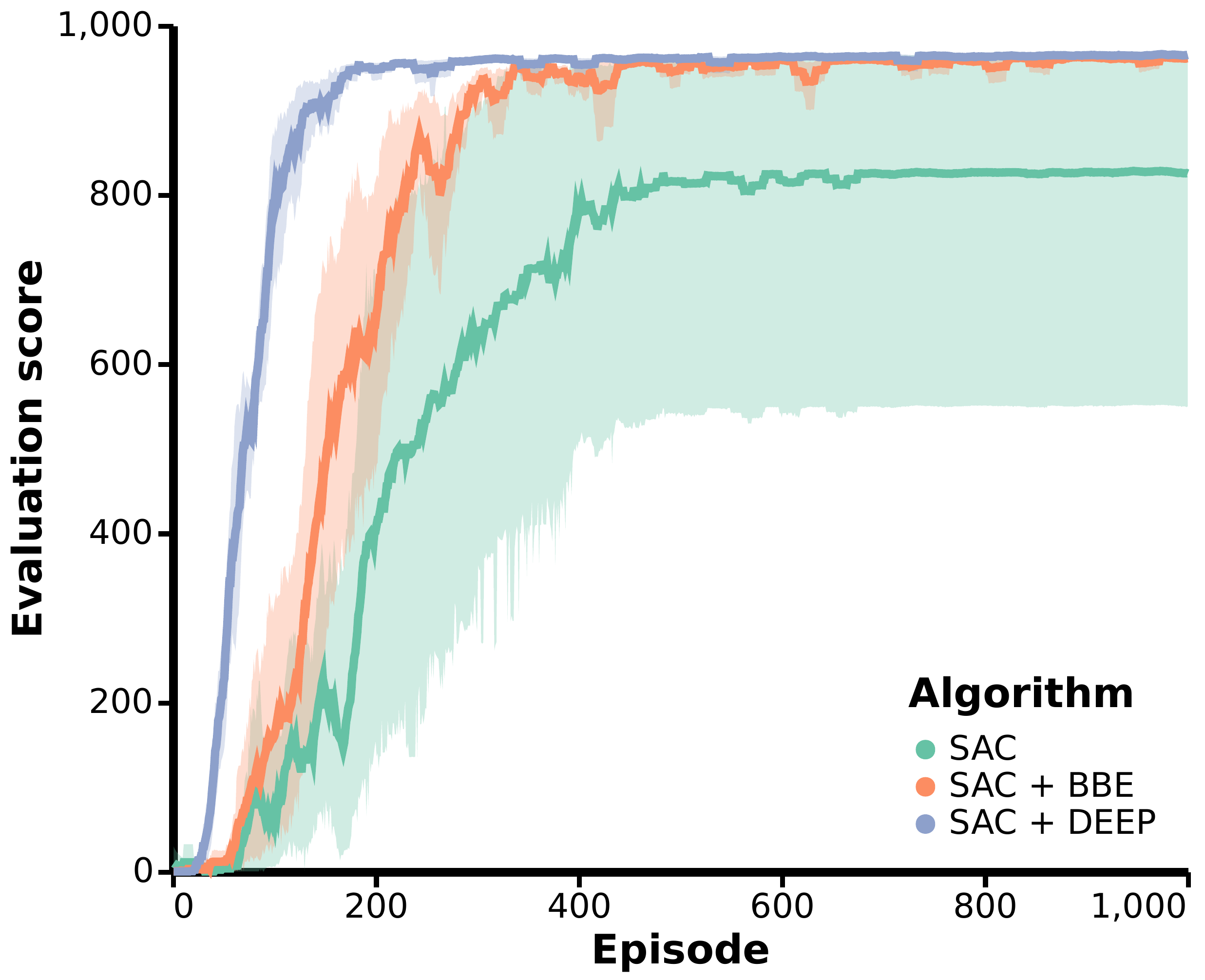}
        \caption{Finger explore}
    \end{subfigure}
    \hfill
    \begin{subfigure}[b]{0.24\textwidth}
        \centering
        \includegraphics[width=\textwidth]{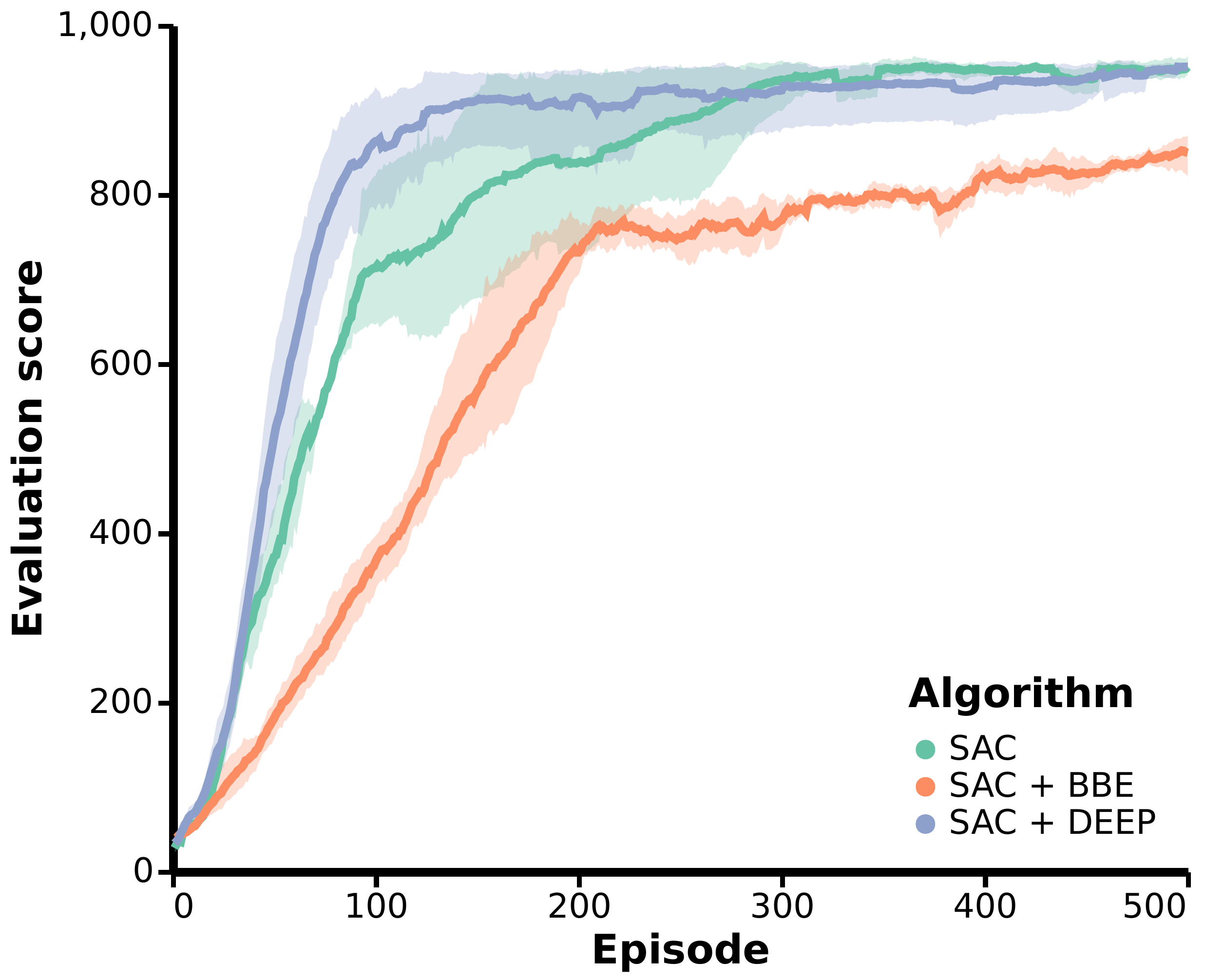}
        \caption{Walker}
    \end{subfigure}
    \begin{subfigure}[b]{0.24\textwidth}
        \centering
        \includegraphics[width=\textwidth]{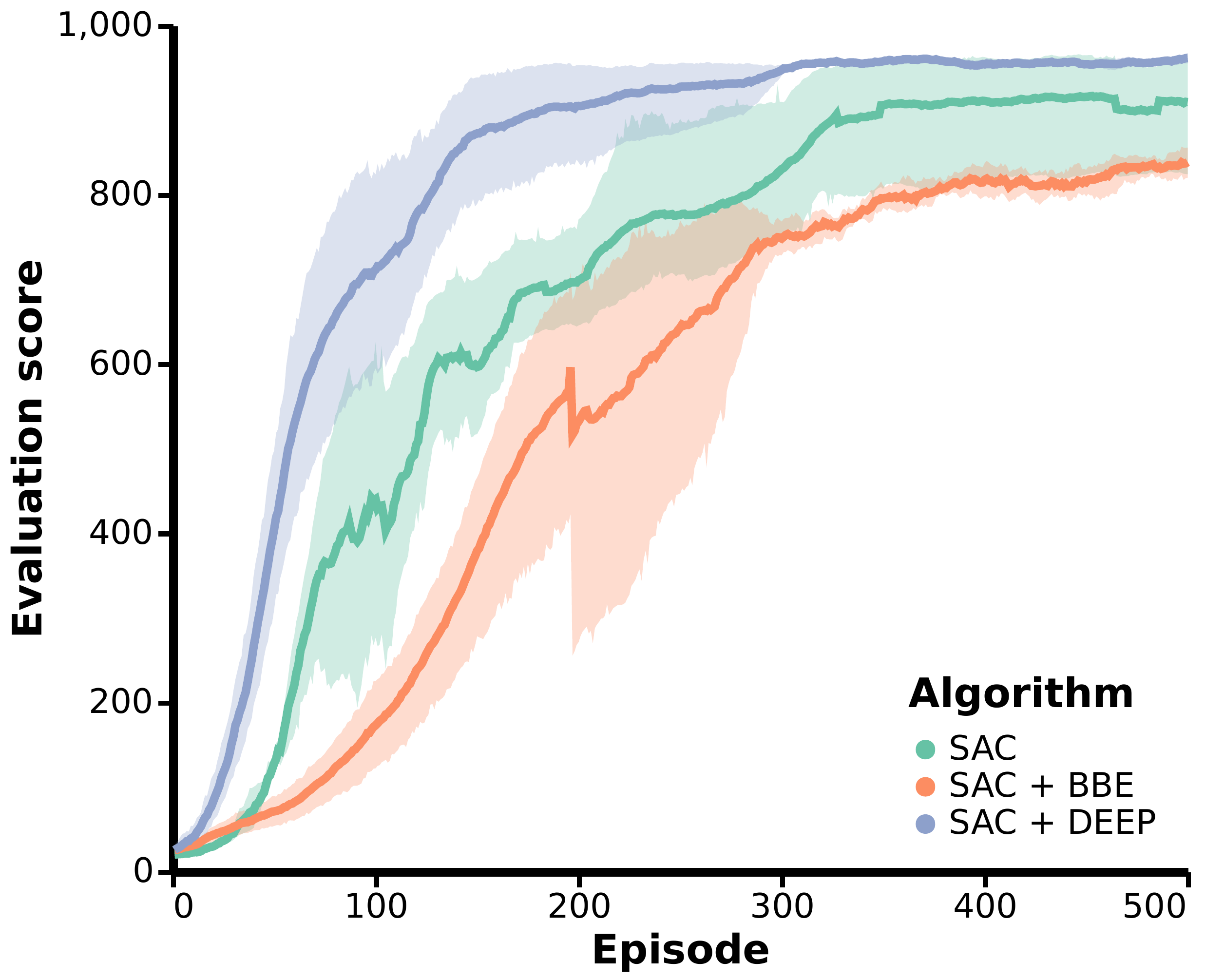}
        \caption{Walker explore}
    \end{subfigure}
    \caption{Results on original Control Suite environments (left in each pair) and modified versions without exploratory resets and rewards (right). 
    Across the original environments, SAC + \algshort{} performs as well or better than SAC, while SAC + BBE performs much worse on some environments.
    On the exploration environments, \algshort{} + SAC learns much faster than SAC.
    BBE sometimes provides significant gains over SAC but sometimes performs worse even on exploration environments.
    }
    \label{fig:control_suite}
\end{figure}

\begin{figure}[b]
    \centering
    \includegraphics[width=0.95\textwidth]{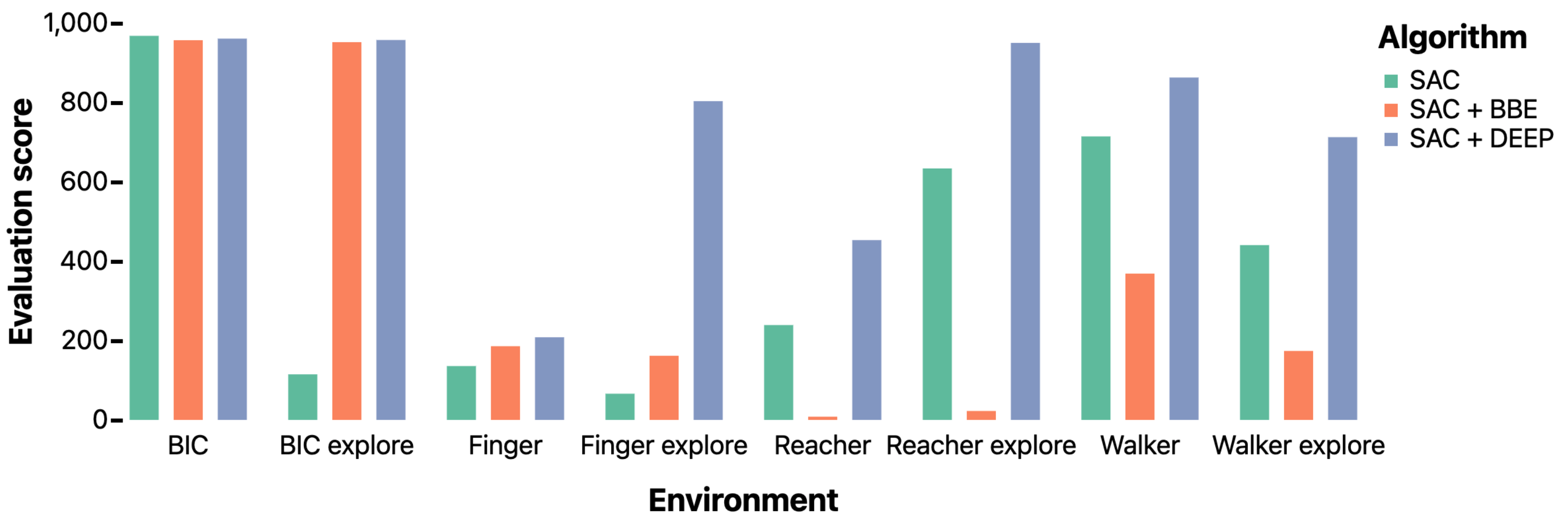}
    \caption{Results after 100 episodes. In this extremely sample-limited regime, exploration speed and fast policy convergence are both essential. In every environment, SAC with \algshort{} (blue, right column in each set of three) performs comparably to or better than SAC alone or SAC with BBE.}
    \label{fig:control_suite_summary}
\end{figure}

We present experiments on the original versions of four Control Suite tasks and their exploration counterparts.
The results are shown in \Cref{fig:control_suite} with the means and 95\% confidence intervals over 8 seeds.
We find that across the original environments, \algshort{} gives similar or slightly better performance to SAC, while BBE significantly impairs SAC on two of the four environments and matches SAC on the other two.
Across the exploration environments, \algshort{} gives the best performance and sample efficiency.
BBE performs better than SAC alone on two exploration environments and worse than SAC on the other two.
\Cref{fig:control_suite_summary} shows the performance of each algorithm after only 100 episodes, highlighting the substantial benefits from using \algshort{} in the few-sample regime.

Overall, SAC + \algshort{} never performs worse than SAC alone, while yielding substantial improvements in environments where rewarding states are harder to discover.
BBE's more mixed performance provides a possible explanation for the limited influence that methods of that family have had on sample-efficient continuous control, and perhaps more generally on sample-efficient RL.
Given that in this setting the addition of BBE is as likely to harm as to help, its lack of adoption is unsurprising.

\section{Related work}

\paragraph{Sample-efficient continuous control.}
Our method leverages progress on sample efficient off-policy RL, as it can be combined with any off-policy algorithm.
A strong line of work has brought the sample complexity of model-free control within range of solving tasks on real robots \citep{Popov2017DataefficientDR,kalashnikov2018qt,haarnoja2018softA,fujimoto2018addressing,haarnoja2018softB,Abdolmaleki2018MaximumAP}.



\paragraph{Bonus-based exploration.}
There have been many bonuses proposed in the BBE framework.
Several works \citep{Stadie2015IncentivizingEI,pathak2017curiosity,burda2018exploration} propose to use prediction error of a learned model to measure a transition's novelty, with the key differences being the state representation used for making predictions. 
\citet{houthooft2016vime} propose a bonus based on the information gain of the policy.
\citet{bellemare2016unifying} and others \citep{ostrovski2017count,Tang2017Exploration} use continuous count analogues to calculate the count-based bonuses of \citet{strehl2008analysis}.
\citet{Machado2020CountBasedEW} use the norm of learned successor features as a bonus, and show that it implicitly counts visits.
Unlike previous work, our paper focuses on the updates and representation of the behavior policy, and \algshort{} can be used in conjunction with any of these bonuses.
Never Give Up \citep{Badia2020NeverGU} uses an episodic exploration bonus and trains policies with different bonus scales including a task policy.
However, it is designed to maximize asymptotic performance rather than sample efficiency and does not learn faster than a baseline early in training.

\paragraph{Optimism.}
Classic exploration methods \citep{Kearns1998NearOptimalRL,Brafman2002RMAXA,strehl2008analysis,Jaksch2008NearoptimalRB}, depend on an optimistically-defined model.
Model-free methods with theoretical guarantees \citep{Strehl2006PACMR,Jin2018IsQP} use optimistically-initialized Q functions.
Similar to our \cref{eq:optimism}, \citet{Rashid2020OptimisticEE} propose a method for ensuring optimism in Q learning with function approximation by using a count function.
However, \algshort{} leaves the task policy unbiased in the few-sample regime by separating the exploration policy from the task policy.

\paragraph{Temporally-extended actions.}
A variety of work proposes to speed up $\epsilon$-greedy exploration via temporally-extended actions which reduce dithering.
Some methods \citep{Schoknecht2003ReinforcementLO,neunert2020continuousdiscrete} propose to bias policies towards repeating primitive actions, resulting in faster exploration without limiting expressivity.
\citet{Dabney2020TemporallyExtendedE} describe a temporally-extended version of $\epsilon$-greedy exploration which samples a random action and a random \emph{duration} for that action.
\citet{Whitney2020DynamicsawareE} use a learned temporally-extended action space representing the reachable states within a fixed number of steps.
While these methods improve over single-step $\epsilon$-greedy, they are unable to perform directed exploration or discover faraway states.


\paragraph{Randomized value functions.}
Modern works \citep{Osband2016DeepEV,osband2019deep} extend Thompson sampling \citep{thompson1933likelihood} to neural networks and the full RL setting.
Relatedly, \citep{Fortunato2018NoisyNF,Plappert2018ParameterSN} learn noisy parameters and sample policies from them for exploration.

\section{Discussion}
In this paper we have investigated the potential for directed exploration to improve the sample efficiency of RL in continuous control.
We found that BBE suffers from bias and slow state coverage, leading to performance which is often worse than undirected exploration.
We introduced \algname{}, which separately learns an unbiased task policy and an exploration policy and combines them to select actions at training time.
\algshort{} pays no performance penalty even on dense-reward tasks and explores faster than BBE.
In our experiments, \algshort{} combined with SAC provides strictly better performance and sample efficiency than SAC alone.
We believe that with its combination of reliable and efficient policy learning across dense and sparse environments, SAC + \algshort{} provides a compelling default algorithm for practitioners.

\subsection*{Acknowledgements}
We thank many people for valuable conversations early in this project, in particular Nicolas Heess, Pablo Sprechmann, Michael Neunert, Jonas Degrave, Jan Humplik, David Abel, Alessandro Ialongo, and Giambattista Parascandolo.

\newpage
\bibliography{bibliography}
\bibliographystyle{bibliography_style}

\newpage

\begin{appendices}

\section{Grid-world visualizations} \label{sec:gridworld-vis}

\begin{figure}[h]
    \centering
    \begin{subfigure}[b]{\textwidth}
        \centering
        \includegraphics[height=0.15\textwidth]{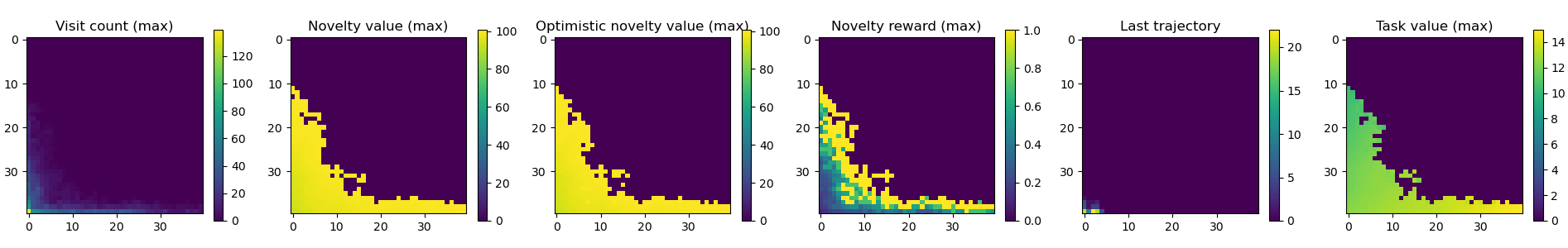}
        \caption{DDQN}
    \end{subfigure}
    
    \begin{subfigure}[b]{\textwidth}
        \centering
        \includegraphics[height=0.15\textwidth]{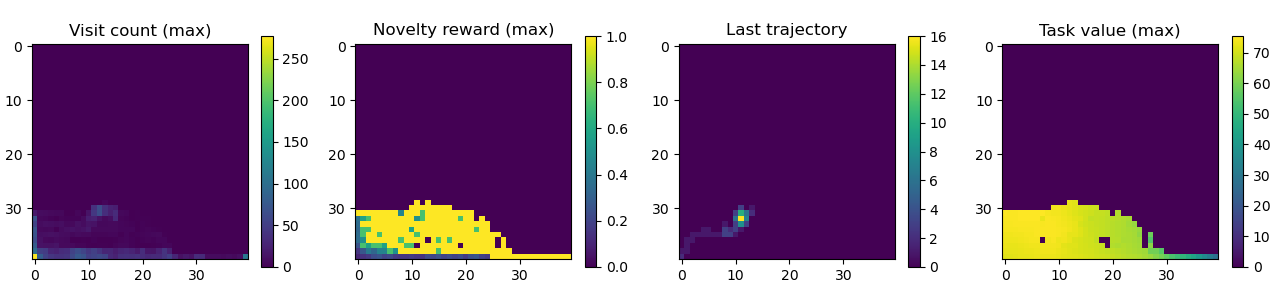}
        \caption{BBE}
    \end{subfigure}
    
    \begin{subfigure}[b]{\textwidth}
        \centering
        \includegraphics[height=0.15\textwidth]{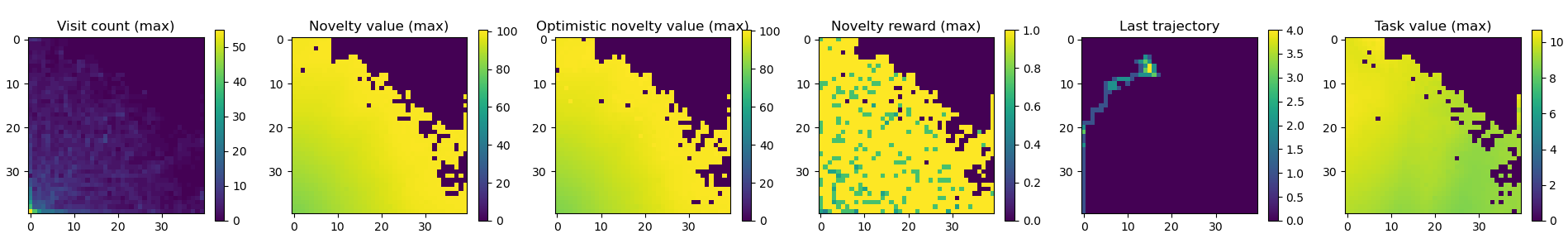}
        \caption{DEEP}
    \end{subfigure}
    \caption{The state of each algorithm after 100 episodes in the grid-world environment.
     Note that the novelty reward and novelty value shown for DDQN are for visualization purposes only, as the algorithm does not use them.
     The ``Task value'' shown for BBE is the sum of the task and novelty rewards, which BBE treats as its objective.
     Un-visited states are marked as zero in each plot.
     The annotation (max) indicates that the value shown is the maximum value for any action at that state.
     ``Last trajectory'' shows the states visited in the last training episode.
     Figures generated by sampling.}
     \label{fig:gridworld_algorithm_states}
\end{figure}

The grid-world environment we use consists of a 40x40 environment with actions up, down, left, and right, with a single start state in the lower left and a single goal state in the upper right.
In \Cref{fig:gridworld_algorithm_states} we show the state of each algorithm after training on the grid-world for 100 episodes.
While DDQN and BBE have only visited a small fraction of the states, \algshort{} has covered most of the environment and will soon solve it.
As they continue to run BBE will eventually find the goal while DDQN will not.
Note that as the figures are rendered by sampling states and actions, some noise in the form of missing states may appear.


\section{Pseudo-count implementation} \label{sec:count_implementation}


Define the Gaussian kernel with dimension $d$ as
\begin{align}
    k_\text{Gauss}(x, x_i) = (2 \pi) ^ {-\frac{d}{2}} \det(\bm{\Sigma})^{-\frac{1}{2}} \exp \left\{-\frac{1}{2} (x - x_i)^\intercal \bm{\Sigma}^{-1} (x - x_i) \right\}.
\end{align}
We can normalize this function to have a maximum at $k(x, x) = 1$ simply by removing the normalizer (everything outside the exponential) by noting that $e^0 = 1$.
This gives the kernel we use:
\begin{align} \label{eq:count_kernel}
    k(x, x_i) = \exp \left\{-\frac{1}{2} (x - x_i)^\intercal \bm{\Sigma}^{-1} (x - x_i) \right\},
\end{align}
where the covariance is a diagonal matrix $\bm{\Sigma} = \text{diag}(\sigma_1^2, \ldots, \sigma_d^2)$. 

To compute $\hat N_n(s, a)$ using this kernel, we perform the following steps:
\begin{enumerate}
    \item Normalize $s$ and $a$:
    \begin{align}
        \bar{s} = \frac{s - \mathcal{S}_\text{min}}{\mathcal{S}_\text{max} - \mathcal{S}_\text{min}} && \bar{a} = \frac{a - \mathcal{A}_\text{min}}{\mathcal{A}_\text{max} - \mathcal{A}_\text{min}}
    \end{align}
    \item Define $x = [\bar{s}, \bar{a}]$ as the concatenation of the normalized state and action.
    \item Compute the kernel from \cref{eq:count_kernel} and sum across all of the previous normalized observations $x_i$:
    \begin{align}
        \hat N_n(s, a) = \hat N_n(x) = \sum_{i=1}^n k(x, x_i)
    \end{align}
\end{enumerate}

For this final step, we leverage the MIT-licensed Kernel Operations library (KeOps, \citet{keops}), a high-performance GPU-accelerated library for efficient computation of reductions on symbolically-defined matrices.
This substantially outperforms implementations in other frameworks, including fully-JITted JAX \citep{jax2018github} version, especially as the dimension of the data grows.

To avoid having to tune the covariance for each environment, we adapt the scaling rule of \citep{Henderson2012NormalRB}.
This rule of thumb requires assumptions on the data (notably, that it comes i.i.d. from a Normal) which are violated in the exploration problem.
However, we find this scaling to be useful in practice.
The rule of thumb bandwidth for dimension $j$ of the data for a multivariate Gaussian kernel is
\begin{align}
    h_j^{ROT} = \left( \frac{4}{2+d}  \right) ^ \frac{1}{4+d} \hat{\sigma}_j n ^ {- \frac{1}{4+d}},
\end{align}
where $\hat{\sigma}_j$ is the empirical variance of dimension $j$ of the data.
Making the assumption that our states are normalized to be in $[-1, 1]$ and distributed uniformly, we can set $\hat{\sigma}_j \approx 0.3$.
As such, in every experiment with continuous-valued states, we set the kernel variance for dimension $j$ as
\begin{align}
    \sigma_j = 0.3 \left( \frac{4}{2+d} \right) ^ \frac{1}{4+d} n ^ {- \frac{1}{4+d}} .
\end{align}
In experiments we find that changes to this scale of less than an order of magnitude make little difference.

Throughout we use 1 for the kernel variance on the action dimensions.

\subsection{Updating the kernel estimator}
Updates to the kernel count estimator consist of appending new normalized observations to the set $\{x_i\}$.
However, we find that computing this kernel becomes prohibitively slow beyond roughly 100K entries, and our experiments run for up to 1M steps.
We take two steps to avoid this slowdown.
Both rely on nonuniform weighting of entries in the observation table when computing the count, leading us to maintain an additional array of weights in addition to the table of observations.

\paragraph{Avoiding duplicate entries.}
If a new observation $x$ has $k(x, x_i) > 0.95$ with some set $M$ of existing indices, we do not add $x$ to the table, and instead add $\nicefrac{1}{|M|}$ to each entry $i \in M$ of the weights table.
In essence, if there is an exact duplicate for $x$ in the table already, we simply count that existing entry twice.
While this is helpful, the probability of observing exact matches decreases rapidly in the dimension of the observations, so this step plays a limited role.

\paragraph{Evicting previous entries.}
Once the length of the observations table reaches some maximum ($2^{15} = 32768$ in our experiments), we evict an existing entry in the table uniformly at random when we make a new entry, thus maintaining the table at that maximum size.
This introduces risk that the exploration bonus would not go to zero in the limit of many environment steps, which we avoid by re-distributing the weight of the evicted observations among those still in the table.
We do this redistribution uniformly; that is, if we evict the entry at location $i$, with weight $w_i$, we add $w_i / (n-1)$ to the weight of each of the $(n-1)$ other entries.
Our reweighting procedure maintains the same \emph{total} amount of count when evicting observations and ensures that bonuses go to zero in the limit of data.
In experiments we find that the exploration rewards earned when using a very small observation table (and thus, many evictions) were practically indistinguishable from using an observation table of unlimited size.

\subsection{Tabular environments}
For the grid-world environment used in \Cref{fig:gridworld_warmstart,fig:gridworld_visits}, we use a tabular visit count rather than pseudo-counts.

\section{Details on rapid Q updates} \label{sec:fast_updates_appendix}

We aim to rapidly update $\qex$ to maximize reward on the non-stationary exploration MDP $\mdp_{R^+_n}$ and thus explore rapidly.
This has three components: (1) updating using the current reward function $R^+_n$ rather than logged rewards, (2) performing many updates to $\qex$ at every timestep using a large learning rate, and (3) using an optimistic version of $\qex$ which is aware of the high value of taking actions that have not yet been explored.
However, aggressively updating $\qex$ poses its own problems; most significantly, Q-learning with function approximation has a tendency to diverge if updated too aggressively with too few new samples.
We use three modifications to the typical Bellman update with target networks \citep{mnih2015human} to mitigate this issue while incorporating optimism.

\begin{itemize}
    \item \textbf{Soft DoubleDQN update.} The DoubleDQN \citep{Hasselt2016DeepRL} update reduces overestimation in Q-learning by selecting and evaluating actions using different parameters.
    We use a soft version of the DoubleDQN update by replacing the max operator with an exponential-Q policy over uniform random actions using a low temperature.
    \item \textbf{Value clipping.} To further mitigate the problem of Q-learning overestimation and divergence, we clip the Bellman targets to be within the range of possible Q values for $\mdp_{R^+_n}$.
    Given that the rewards $r^+$ are scaled to be in $[0, 1]$, any policy would have a value $\qex(s, a) \in [0, \bar{r}]$, where $\bar{r} = \nicefrac{1}{1 - \gamma}$.
    \item \textbf{Optimistic targets.} We use the optimistic adjustment in \cref{eq:optimism} when computing targets.
\end{itemize}

Define the softmax-Q policy for some Q function $Q$ as 
\begin{align}
    \pi(a \mid s; Q) = \frac{\exp \big\{ \nicefrac{Q(s, a)}{\tau} \big\}}{\int_\mathcal{A} \exp \big\{ \nicefrac{Q(s, a')}{\tau}  ~da' \big\}}
\end{align}
which we approximate using self-normalized importance sampling with a uniform proposal distribution. 
The target for updating $\qex$ is
\begin{align} \label{eq:update_target}
    y(s, a, s') = \text{clip} \left( R^+_n(s, a) + \gamma \E_{a' \sim \pi(\cdot \mid s'; \qex^+)} \Big[ \qex^+(s', a'; \theta^-) \Big], 0, \bar{r} \right).
\end{align}
where $R^+_n(s, a)$ is the current exploration bonus, which we recompute at update time; $\qex^+(s', a'; \theta^-)$ is the target network for the exploration value function, with optimism applied.
We then minimize the squared error between $y(s, a, s')$ and $\qex(s, a)$.

\section{Environments for exploration} \label{sec:environments_appendix}

To enable benchmarking the performance of exploration methods on continuous control, we constructed a new set of environments.
Our motivation comes from the challenges of performing resets and defining shaped rewards in real-world robotics, where it is not possible to measure and set states exactly.
Unlike in simulation, it may be difficult or impossible to implement uniform resets of the robot and the objects in the scene; states with a walking robot standing upright or a block in midair require significant expertise to reach.
Similarly many shaped rewards in simulation rely on precise knowledge of the locations of objects in a scene to provide rewards corresponding to e.g. an objects distance from a goal.
We make modifications which capture the spirit of these real-world constraints, though these exact environments might still be difficult to construct in the real world:
\begin{itemize}
    \item Small reset distributions. 
    Instead of resetting every object in the scene uniformly in the space, we randomize each object's configuration over a smaller set of starting states.
    This reflects some properties of real environments, such as walking robots starting on the ground instead of midair, or the object in a manipulation task not starting in its goal receptacle.
    \item Sparse rewards.
    While dense rewards are difficult to construct without real-time monitoring of object positions, sparse rewards are often simpler.
    A picking task, for example, can provide a sparse reward simply by checking whether the desired object is inside a receptacle.
\end{itemize}

As the base environments for our benchmark, we select four tasks from DeepMind Control Suite \citep{tassa2018deepmind}, an Apache-licensed standard set of benchmarks implemented using the commercial MuJoCo physics simulator \citep{todorov2012mujoco}.
Denoting the state (observation) and action dimensions of an environment as $\text{dim}(\mathcal{S}) \rightarrow \text{dim}(\mathcal{A})$, these environments are:
\begin{itemize}
    \item \textbf{Ball-in-cup catch} (manipulation, $8d \rightarrow 2d$).
    \item \textbf{Reacher hard} (goal-directed, $6d \rightarrow 2d$).
    \item \textbf{Finger turn\_hard} (manipulation, goal-directed, $12d \rightarrow 2d$).
    \item \textbf{Walker walk} (locomotion, $24d \rightarrow 6d$).
\end{itemize}

We modify each environment to remove the accommodations of wide reset distributions and sparse rewards which make them easy to solve without directed exploration. 
The new environments and their changes are as follows:
\begin{itemize}
    \item \textbf{Ball-in-cup explore} (manipulation, $8d \rightarrow 2d$). The original task resets the ball uniformly in the reachable space, including already in the cup (the goal state). 
    We modify the environment to only reset the ball in a region below the cup, as if the ball was hanging and gently swinging.
    The original task already has sparse rewards.
    \item \textbf{Reacher explore} (goal-directed, $6d \rightarrow 2d$).
    The original task samples arm positions and goal positions uniformly, resulting in the arm being reset very near the goal.
    We modify the reset distribution to only include states with the arm mostly extended to the right and targets opposite it on the left in a cone with angle $\pi/2$.
    Note that this task is somewhat easier than the original since the policy only needs to navigate between smaller regions of the space, but is harder due to the resets not providing exploration.
    The original task already has sparse rewards.
    \item \textbf{Finger explore} (manipulation, goal-directed, $12d \rightarrow 2d$).
    The original task resets the finger uniformly, the angle of the spinner uniformly, and the goal location uniformly on the circle reachable by the tip of the spinner.
    We modify the environment to reset the finger joints each in the quadrant pointing away from the spinner, the spinner pointing in the downward quadrant, and the target in the upward quadrant.
    Similarly to Reacher explore, this task is simpler than the original but harder to explore in.
    The original task already has sparse rewards.
    \item \textbf{Walker explore} (locomotion, $24d \rightarrow 6d$).
    The original environment resets the walker just above the ground with random joint angles, leading to it frequently starting upright and in position to begin walking.
    We modify the environment by allowing time to progress for 200 steps in the underlying physics (20 environment steps), which is enough time for the walker to fall to the floor.
    This simulates the walker starting in a random lying-down configuration.
    The original rewards include a sparse reward for being upright and above a certain height and a linear component for forward velocity.
    We replace the forward velocity reward with a sparse version which provides reward only when the agent is moving at or above the target speed.
\end{itemize}


\section{Experimental implementation details} \label{sec:appendix_benchmark_implementation}

\subsection{Computing infrastructure}
We implemented \algshort{} using JAX \citep{jax2018github} and the neural network library Flax \citep{flax2020github} which is built on it, both of which are Apache-licensed libraries released by Google.
The SAC implementation we use is from \citet{pytorch_sac} (MIT-licensed), built on Pytorch \citep{Paszke2019PyTorchAI} (custom BSD-style license).
The experiments in this paper take about a week to run using 16 late-model NVidia GPUs by running 2-4 seeds of each experiment at once on each GPU.

\subsection{Network architectures and training}
For the Q networks used as $\qex$ in continuous environments and as the task policy in the grid-world experiments, we use fully-connected networks with two hidden layers of 512 units each and ReLU activations.
These networks flatten and concatenate the state and action together to use as input and produce a 1d value prediction.
This enables us to use the same networks and training for discrete and continuous actions rather than using the usual discrete-action trick of simultaneously producing a Q-value estimate for every action given the state.

These Q networks are trained using the Adam optimizer \citep{kingma2014adam} with learning rate $10^{-3}$.
$\qex$ is updated with two Bellman updates with batch size 128 per environment step.
We update the target network after every environment step for $\qex$ to allow very rapid information propagation about changing rewards.
For the Q network defining $\pitask$ in the grid-world we use a learning rate of $10^{-4}$ and update the target network after every 50 Bellman updates.

\subsection{Other hyperparameters}
We draw 64 samples from $\pitask$ when computing the behavior policy and 64 samples from a uniform distribution over actions when updating $\qex$ as described in Appendix \Cref{sec:fast_updates_appendix}.
We set the temperature for Boltzmann sampling from all Q-network policies as $\tau = 0.1$.
$\qex$ uses a discount $\gamma = 0.99$.

\newpage
\section{Additional benchmark results} \label{sec:benchmark_results_appendix}

We performed an experiment to check whether it was simply the separation of learning two separate Q functions which enabled \algshort{}'s performance.
To do this, we modified SAC + BBE to learn one Q function for the task reward function and one Q function for the exploration reward function.
The policy was then trained to maximize the sum of those two Q functions.
This baseline, which we call SAC 2Q, performed uniformly worse than SAC + BBE, but we include its results here for completeness.


\begin{figure}[h]
    \centering
    \begin{subfigure}[b]{0.24\textwidth}
        \centering
        \includegraphics[width=\textwidth]{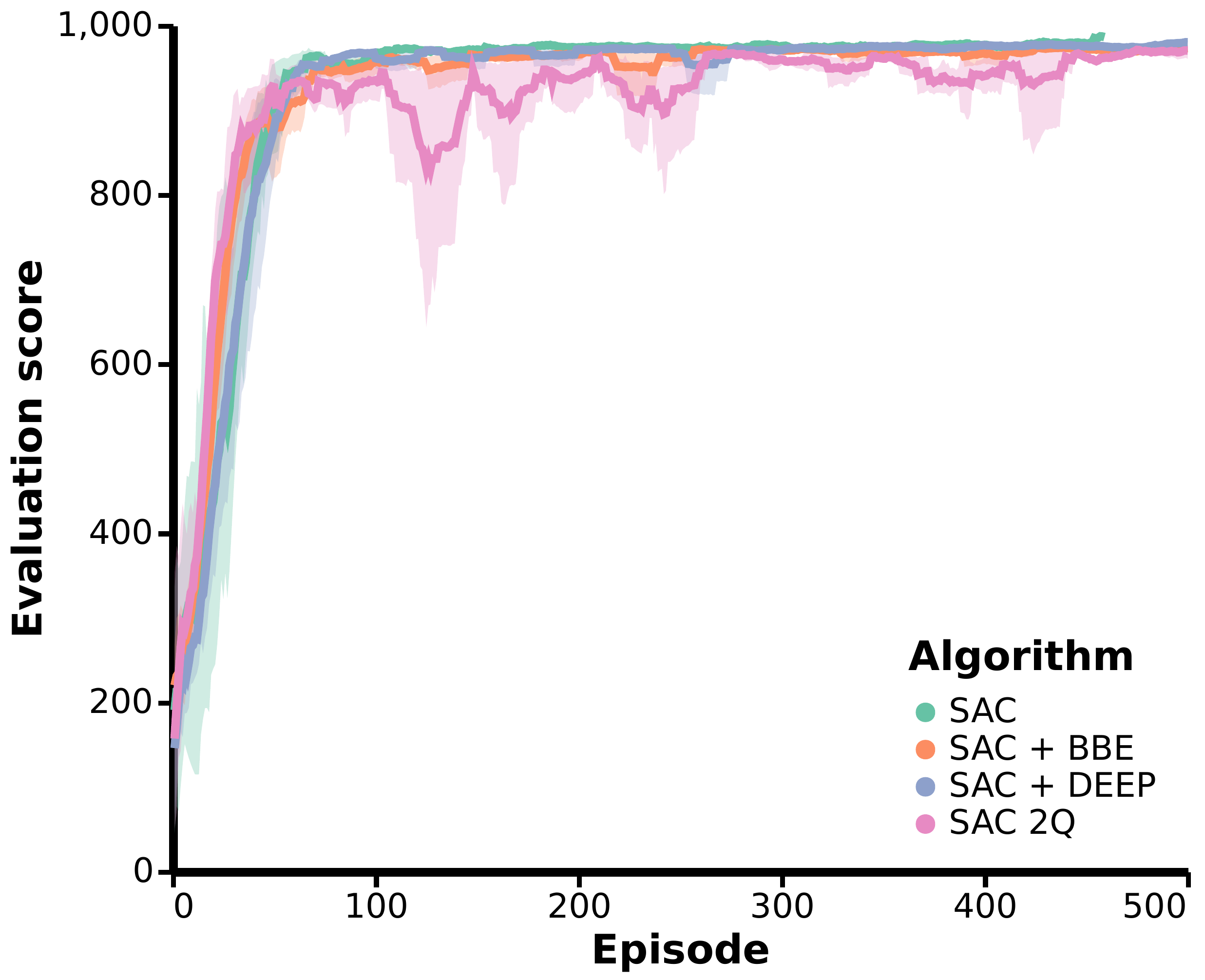}
        \caption{Ball-in-cup}
    \end{subfigure}
    \begin{subfigure}[b]{0.24\textwidth}
        \centering
        \includegraphics[width=\textwidth]{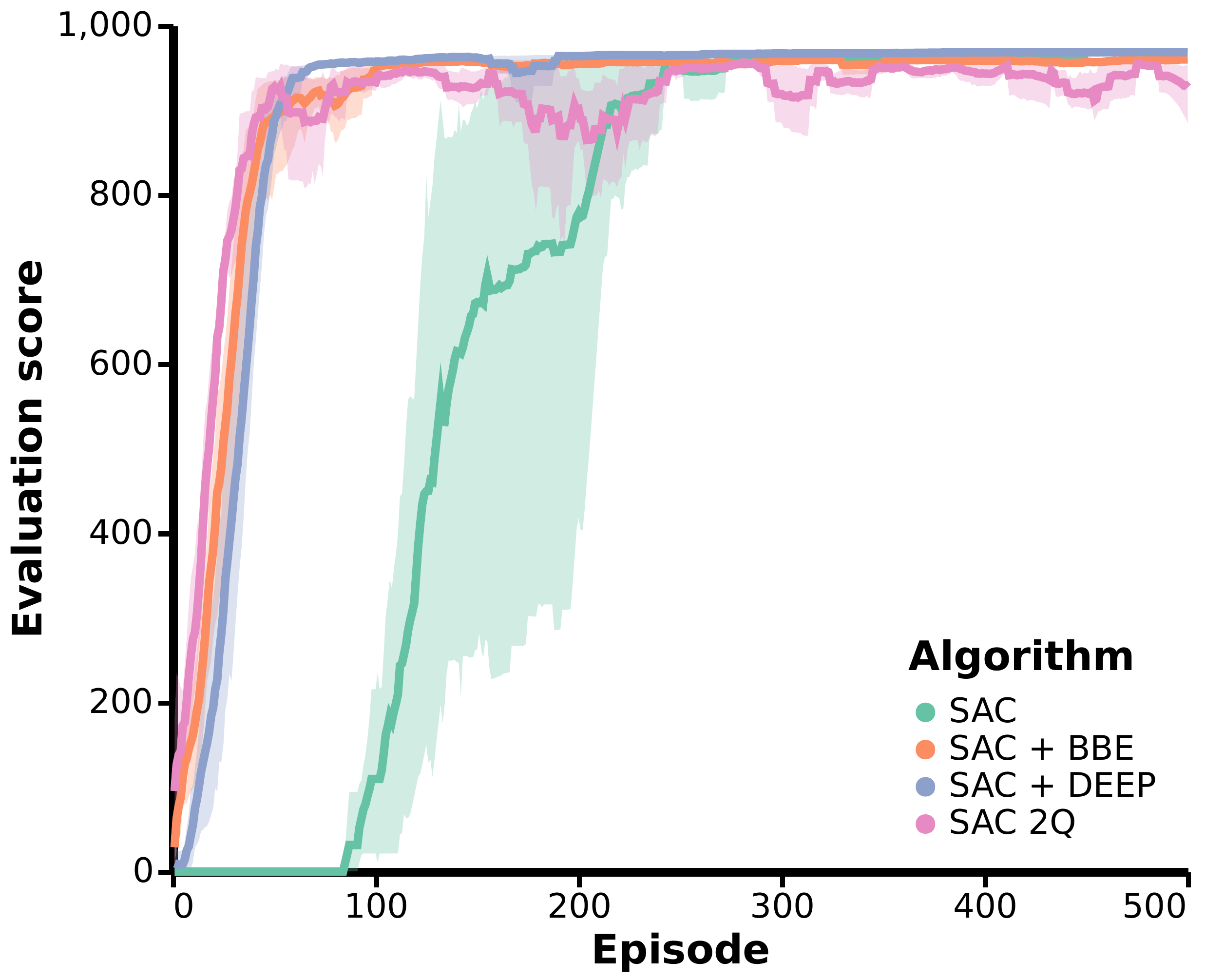}
        \caption{Ball-in-cup explore}
    \end{subfigure}
    \hfill
    \begin{subfigure}[b]{0.24\textwidth}
        \centering
        \includegraphics[width=\textwidth]{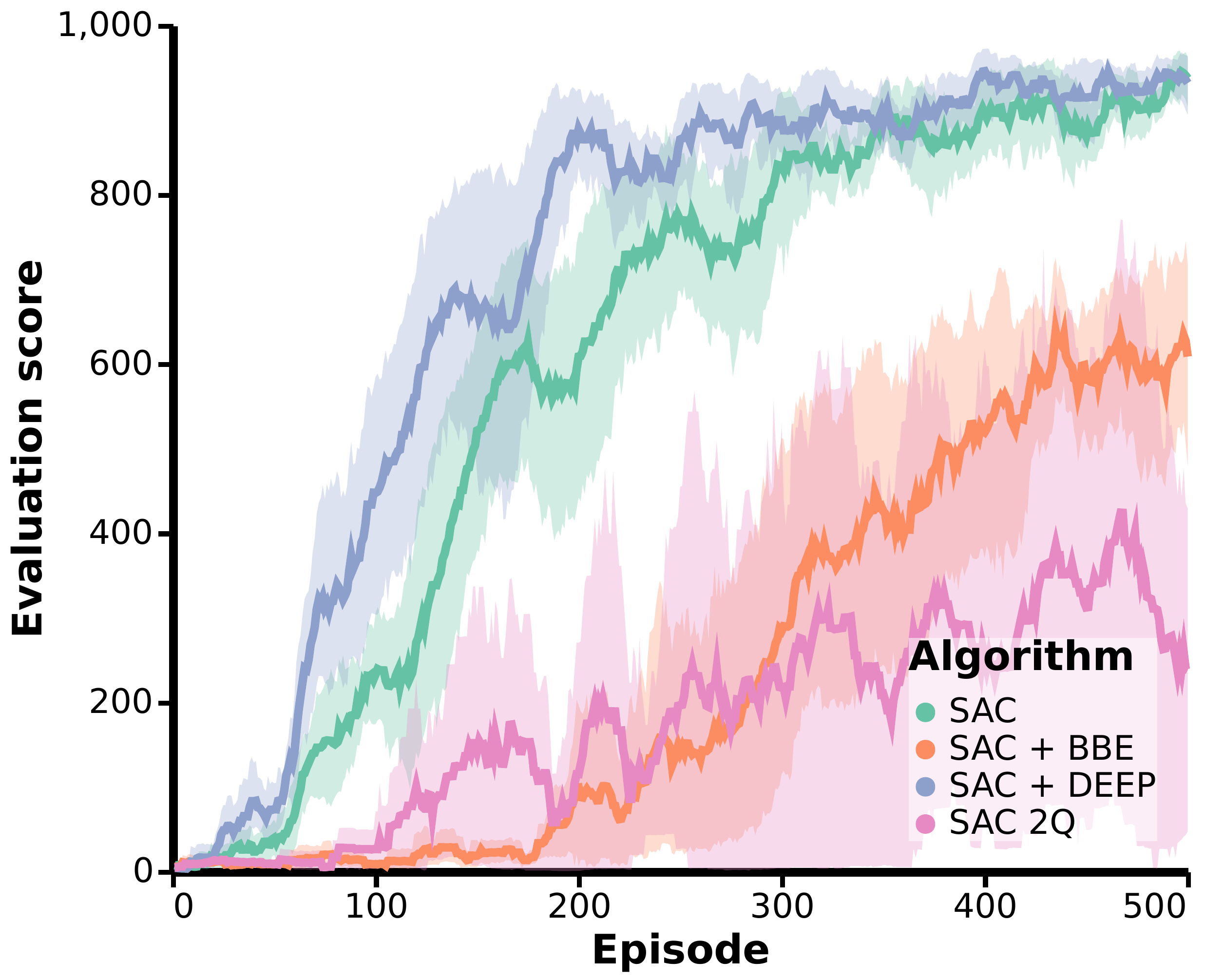}
        \caption{Reacher}
    \end{subfigure}
    \begin{subfigure}[b]{0.24\textwidth}
        \centering
        \includegraphics[width=\textwidth]{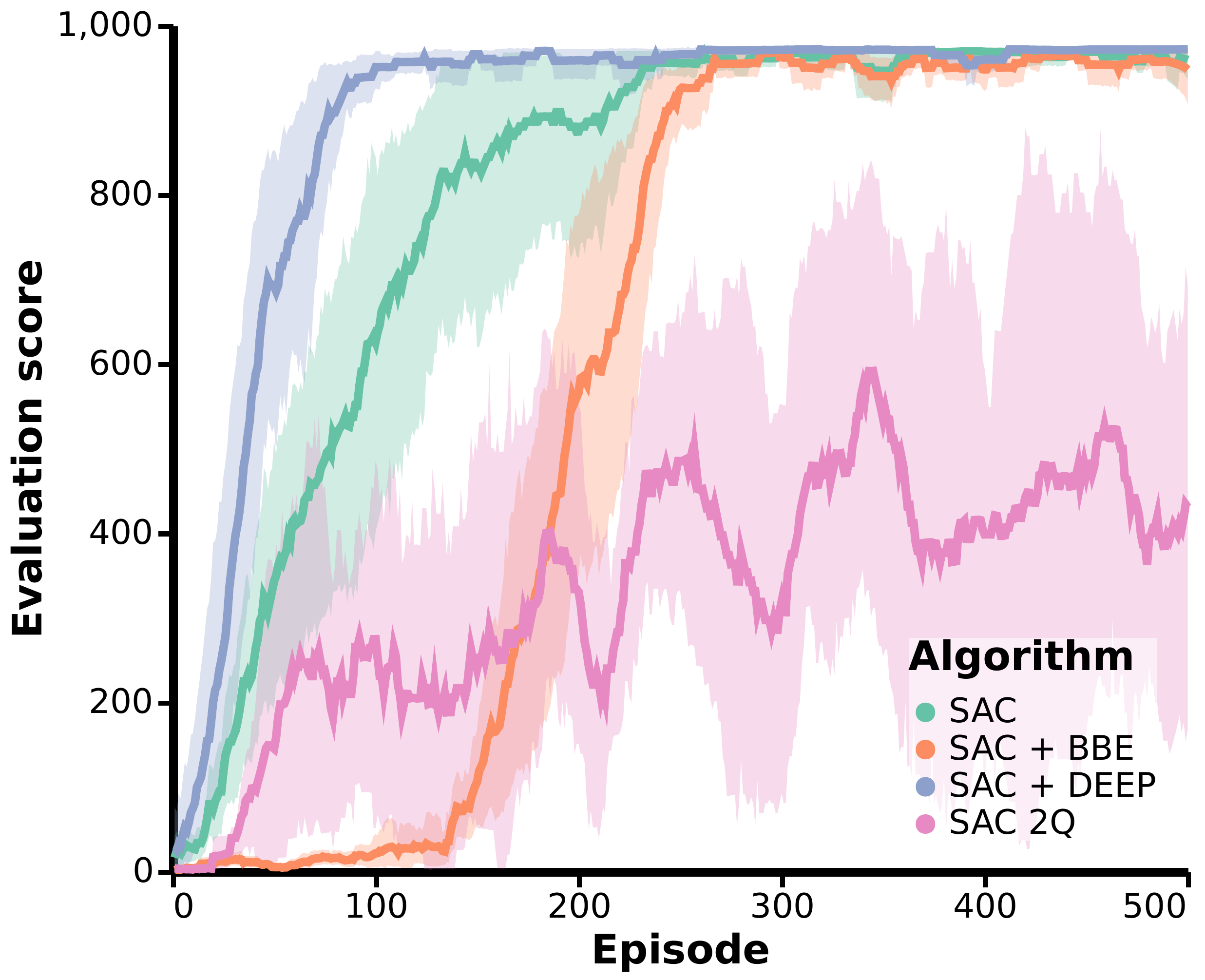}
        \caption{Reacher explore}
    \end{subfigure}
    \vspace{1em}

    \begin{subfigure}[b]{0.24\textwidth}
        \centering
        \includegraphics[width=\textwidth]{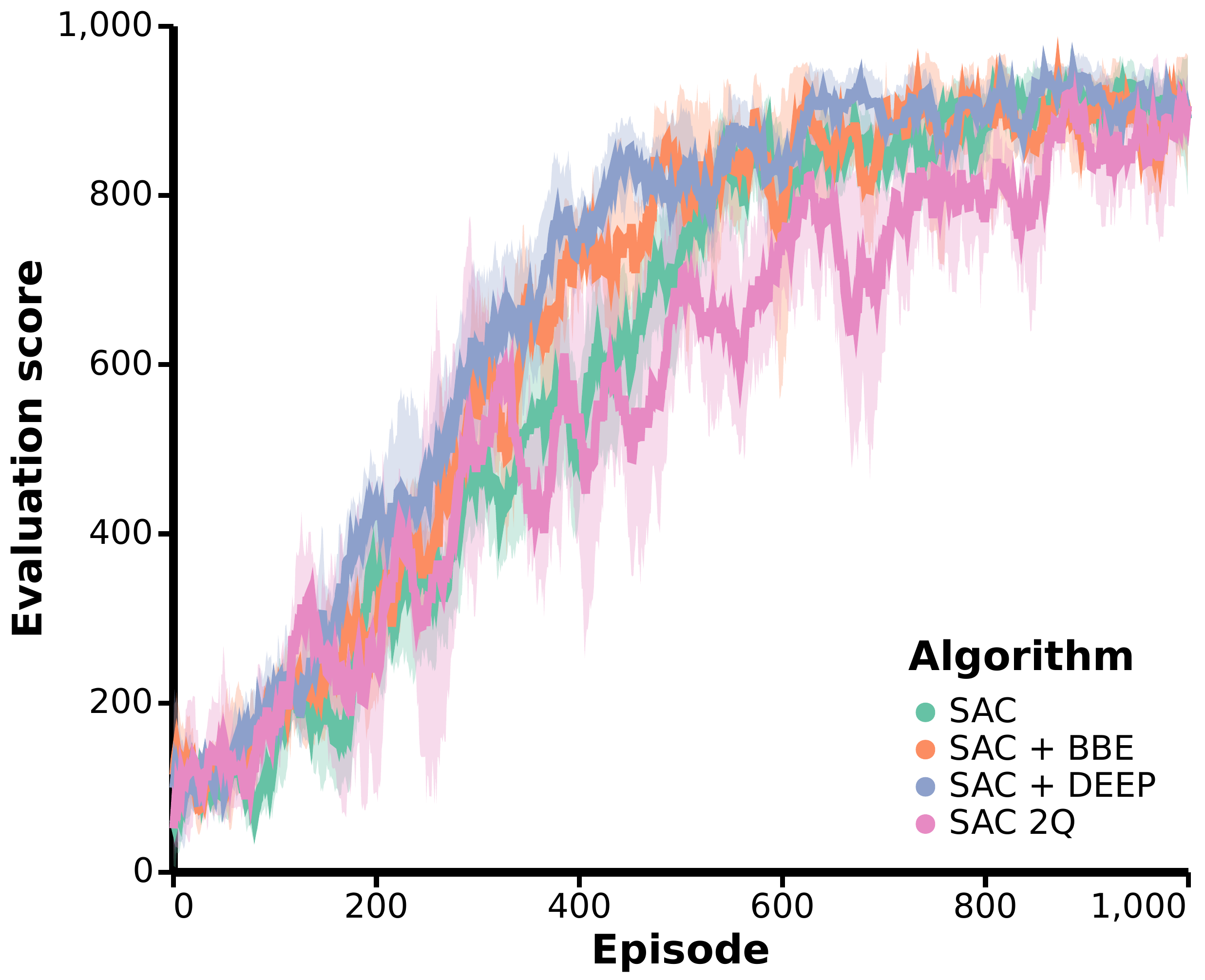}
        \caption{Finger}
    \end{subfigure}
    \begin{subfigure}[b]{0.24\textwidth}
        \centering
        \includegraphics[width=\textwidth]{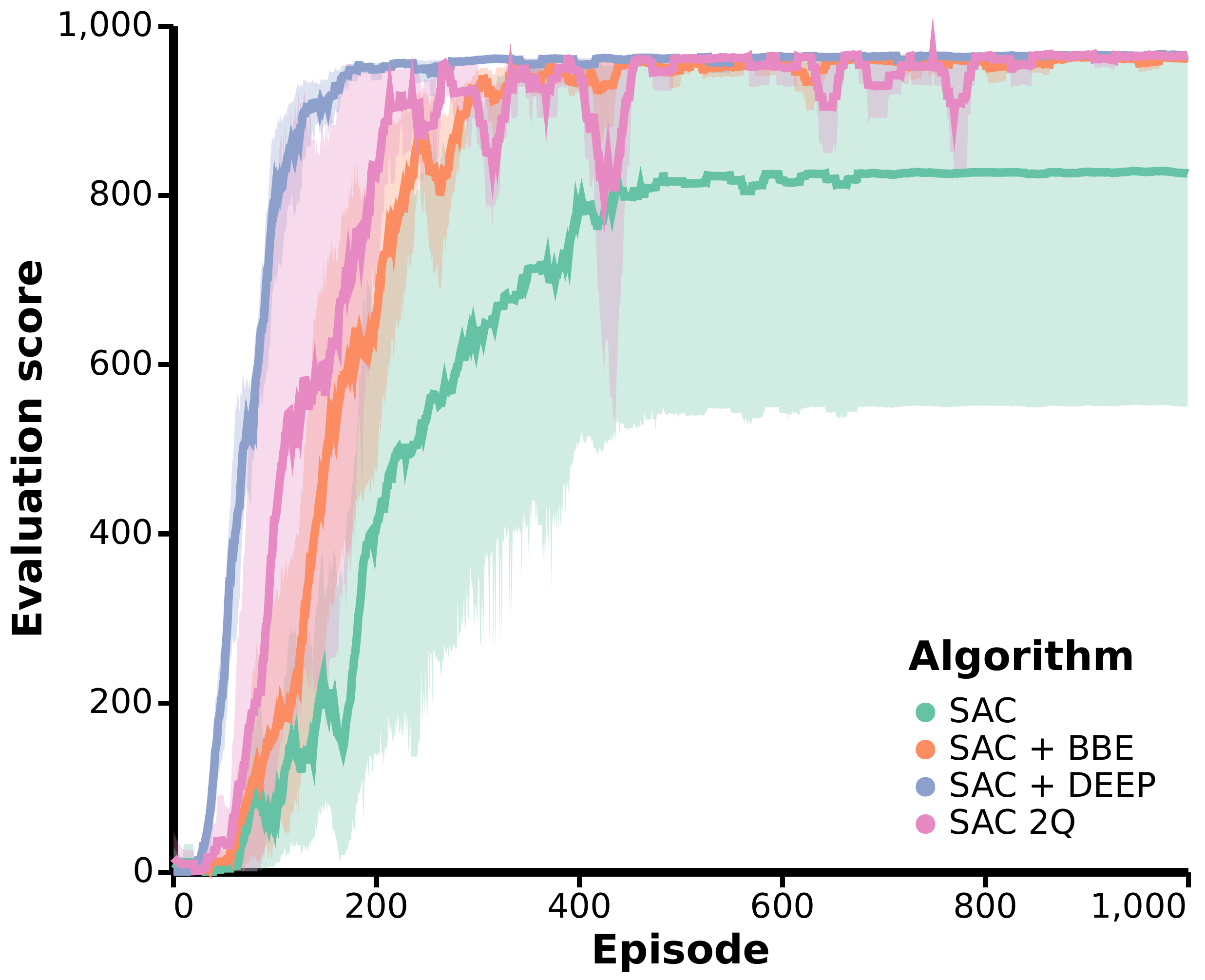}
        \caption{Finger explore}
    \end{subfigure}
    \hfill
    \begin{subfigure}[b]{0.24\textwidth}
        \centering
        \includegraphics[width=\textwidth]{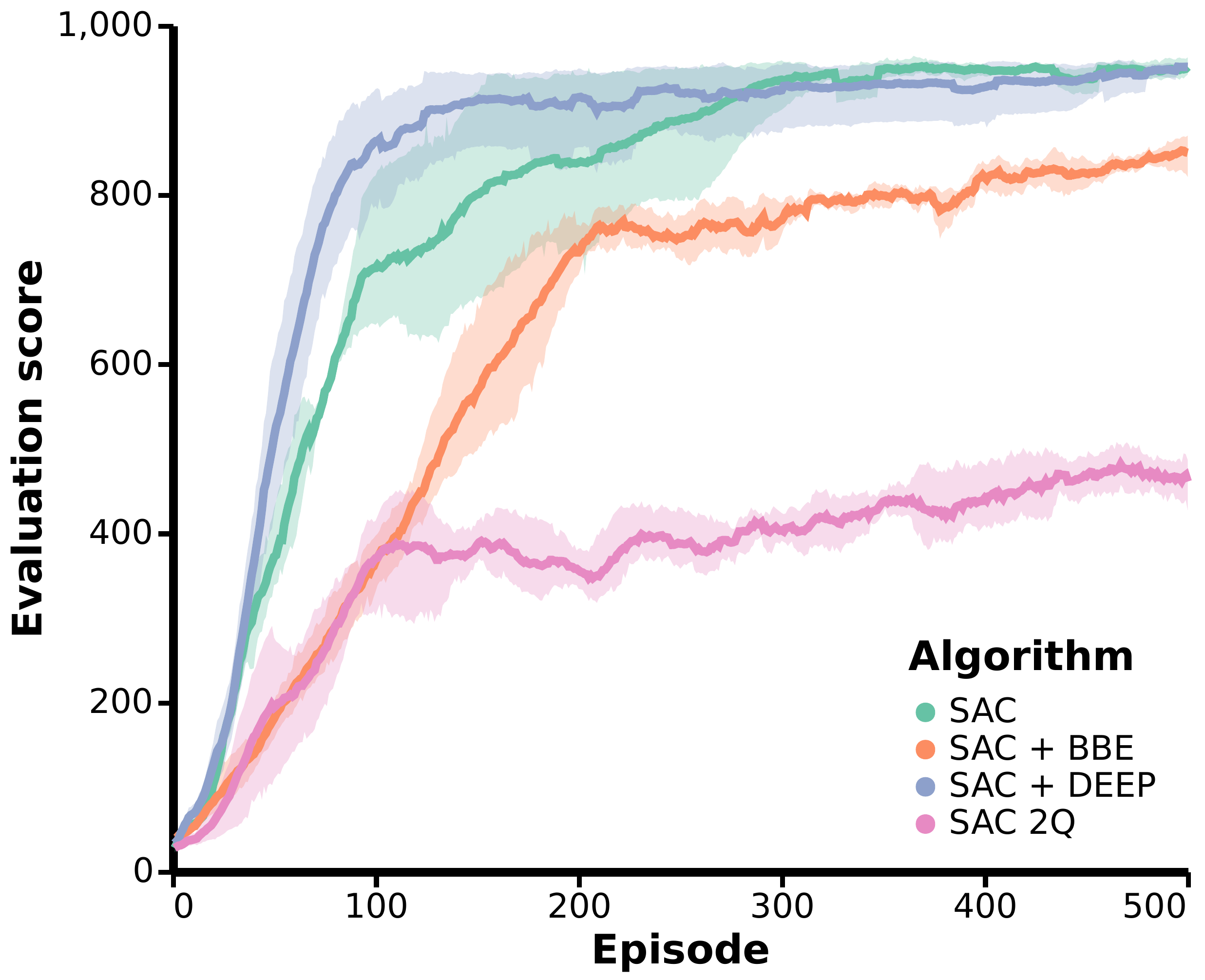}
        \caption{Walker}
    \end{subfigure}
    \begin{subfigure}[b]{0.24\textwidth}
        \centering
        \includegraphics[width=\textwidth]{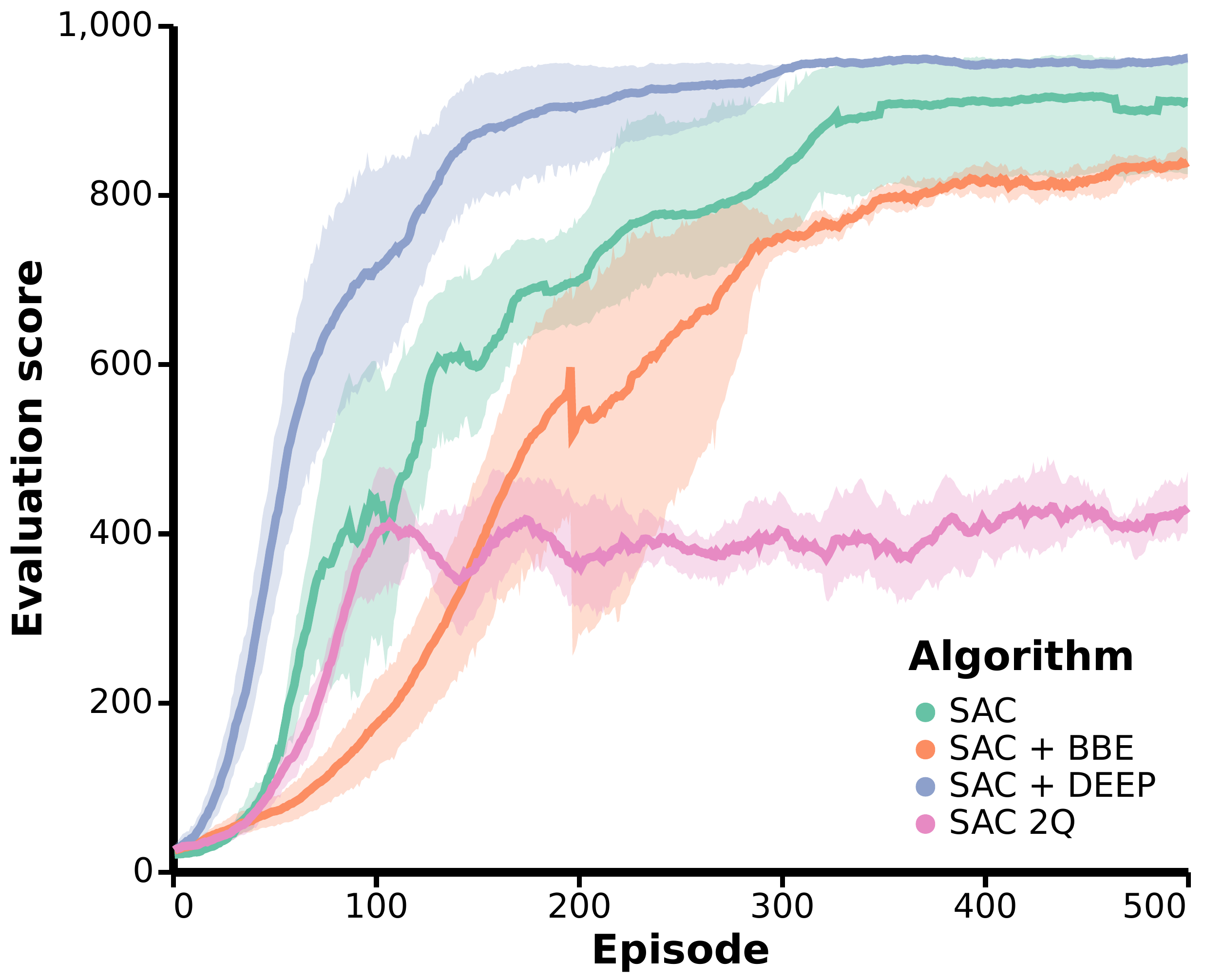}
        \caption{Walker explore}
    \end{subfigure}
    \caption{SAC 2Q performs uniformly worse than SAC + BBE.}
\end{figure}


\end{appendices}
\end{document}